\PassOptionsToPackage{table,dvipsnames}{xcolor}
\documentclass[sigconf,screen]{acmart}
\AtBeginDocument{%
  }

\copyrightyear{2026}
\acmYear{2026}
\setcopyright{cc}
\setcctype{by}
\acmConference[CAIS '26]{ACM Conference on AI and Agentic Systems}{May 26--29, 2026}{San Jose, CA, USA}
\acmBooktitle{ACM Conference on AI and Agentic Systems (CAIS '26), May 26--29, 2026, San Jose, CA, USA}
\acmDOI{10.1145/3786335.3813129}
\acmISBN{979-8-4007-2415-2/2026/05}

\acmSubmissionID{52}

\usepackage{amsfonts}
\usepackage{amsmath}
\usepackage{booktabs}
\usepackage{caption}
\usepackage{hyperref}
\usepackage{listings}
\usepackage{multirow}
\usepackage{subcaption}

\newcommand{\poscolor}[1]{{\color{ForestGreen}#1}}
\newcommand{\negcolor}[1]{{\color{BrickRed}#1}}

\definecolor{roleColor}{RGB}{0, 51, 102}    %
\definecolor{placeColor}{RGB}{153, 0, 76}   %

\lstdefinestyle{booktabsstyle}{
    basicstyle=\ttfamily\small,
    breaklines=true,
    breakatwhitespace=true,
    columns=fullflexible,
    keepspaces=true,
    backgroundcolor=\color{gray!10},
    frame=tb, 
    rulecolor=\color{black},
    framesep=10pt,
    breakindent=0pt, %
    framexrightmargin=10pt,
    framexleftmargin=10pt, 
    float=t,
    captionpos=b,
    abovecaptionskip=10pt,
    moredelim=[s][\color{placeColor}]{\{}{\}},
    literate={%
        {[system]}{{\textcolor{roleColor}{\textbf{[system]}}}}1
        {[user]}{{\textcolor{roleColor}{\textbf{[user]}}}}1
    }
}

\begin{document}

\title[Do Agents Need to Plan Step-by-step?]{\texorpdfstring{Do Agents Need to Plan Step-by-Step?\\Rethinking Planning Horizon in Data-Centric Tool Calling}{Do Agents Need to Plan Step-by-Step? - Rethinking Planning Horizon in Data-Centric Tool Calling}}

\author{Naoki Otani}
\email{naoki@megagon.ai}
\affiliation{%
  \institution{Megagon Labs}
  \city{Mountain View}
  \state{California}
  \country{USA}
}

\author{Nikita Bhutani}
\email{nikita@megagon.ai}
\affiliation{%
  \institution{Megagon Labs}
  \city{Mountain View}
  \state{California}
  \country{USA}
}

\author{Hannah Kim}
\email{hannah@megagon.ai}
\affiliation{%
  \institution{Megagon Labs}
  \city{Mountain View}
  \state{California}
  \country{USA}
}

\author{Dan Zhang}
\email{dan_z@megagon.ai}
\affiliation{%
  \institution{Megagon Labs}
  \city{Mountain View}
  \state{California}
  \country{USA}
}

\author{Estevam Hruschka}
\email{estevam@megagon.ai}
\affiliation{%
  \institution{Megagon Labs}
  \city{Mountain View}
  \state{California}
  \country{USA}
}

\renewcommand{\shortauthors}{Otani et al.}

\begin{abstract}

Explicit planning is a critical capability for LLM-based agents solving complex data-centric tasks, which require precise tool calling over external data sources. Existing strategies fall into two paradigms based on planning horizon: (1) full-horizon (FH), which generates a complete plan before execution, and (2) single-step horizon (SH), which interleaves each action (tool call) with incremental reasoning and observation. While step-by-step execution is a common default under the assumption that \textit{eager} execution monitoring is necessary for adaptability, we revisit this assumption for well-defined data-centric tasks. Our controlled empirical study isolates planning horizon as the key architectural feature and systematically analyzes the effects of topological complexity and tool robustness on both paradigms. Our experiments across Knowledge Base Question Answering and Multi-hop QA show that FH planning with \textit{lazy} replanning achieves accuracy parity with SH across varying depths, breadths, and robustness levels, while using $2$--$3\times$ fewer tokens. These findings suggest that for well-defined data-centric tasks, eager step-wise monitoring is often unnecessary, and full-horizon planning with on-demand replanning can offer a more efficient default.

\end{abstract}



\maketitle

\section{Introduction}\label{sec:introduction}

\begin{figure}[t]
    \centering
    \begin{minipage}{0.9\linewidth}
        \centering
        \includegraphics[width=1\linewidth]{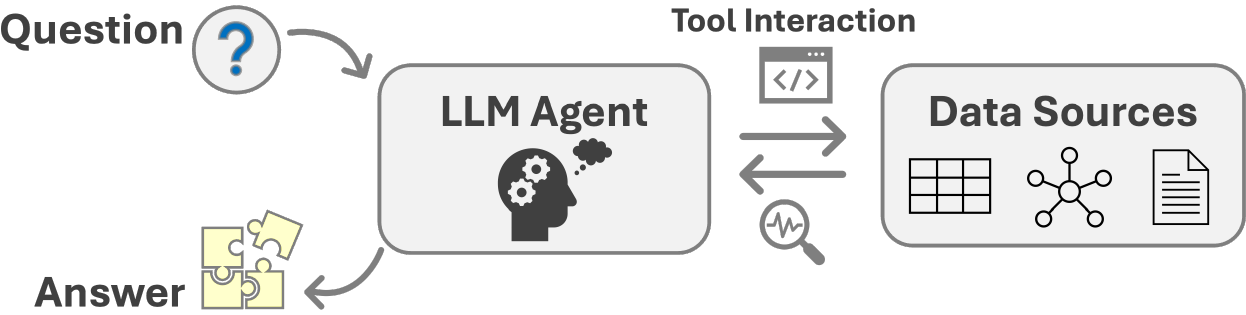}
        \subcaption{In data-centric tasks, LLM agents must coordinate tool calls to synthesize the answer from external data sources. This complex tool calling necessitates explicit planning (below).}
        \label{fig:data-centric-tasks}
    \end{minipage}\\
    \vspace{4pt}
    \begin{minipage}{1\linewidth}
        \centering
        \includegraphics[width=1\linewidth]{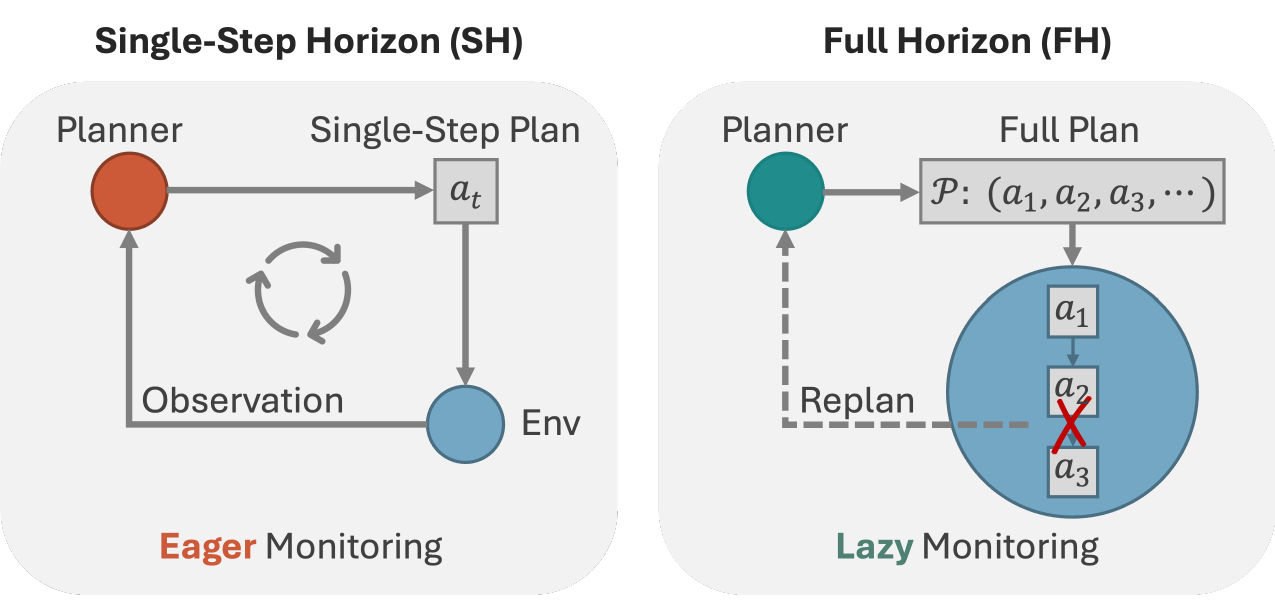}
        \subcaption{Single-step horizon (SH) alternates planning and execution at each step (\emph{eager} monitoring). Full-horizon (FH) plans upfront and (optionally) replans only on demand (\emph{lazy} monitoring).}
        \label{fig:introduction-pne-vs-itr}
    \end{minipage}\\
    \vspace{4pt}
    \begin{minipage}{1\linewidth}
        \centering
        \includegraphics[width=1\linewidth]{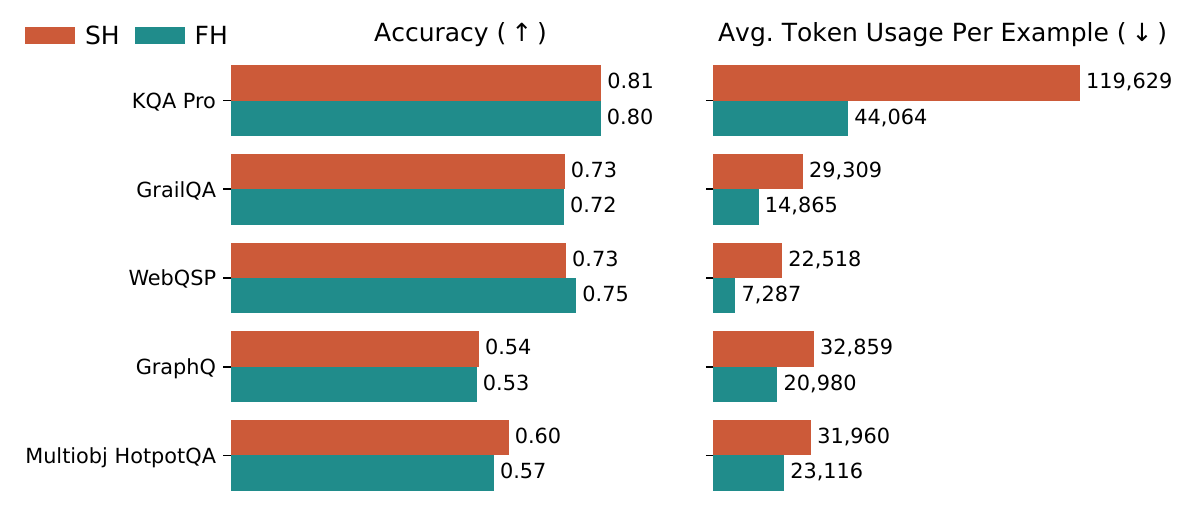}
        \subcaption{When GPT-4.1-mini is used as a backbone LLM, SH and FH achieve comparable accuracy across datasets (left), but SH consumes much more input+output tokens (right). See \S\ref{sec:experiments} for details.}
        \label{fig:introduction-comparison}
    \end{minipage}
    \caption{\textbf{Planning horizon in data-centric tool calling.} SH plans and executes step-by-step. FH plans ahead and replans only when needed, which can reduce token consumption.}
    \Description{A three-part overview of planning horizon in data-centric tool calling. The top panel shows that answering data-centric questions requires coordinating multiple tool calls over external sources. The middle panel contrasts single-step horizon planning, which alternates planning and execution after each tool call, with full-horizon planning, which generates a multi-step tool-use plan upfront and replans only when needed. The bottom panel summarizes experimental results showing similar accuracy between the two approaches across datasets, while single-step horizon uses substantially more tokens.}
    \label{fig:introduction}
\end{figure}

Large Language Model (LLM) agents are increasingly deployed to solve \textbf{data-centric tasks} in which answers must be constructed through tool calls over external sources such as databases, knowledge graphs, or documents~(Figure~\ref{fig:data-centric-tasks}). In these settings, success depends on coordinating tool calls that match the latent logic (e.g., joins or multi-hop reasoning) and vocabulary constraints imposed by the data source (e.g., data schema, entity mentions). As these tasks grow in complexity, explicit planning of low-level tool calls has become a central component of modern agentic architectures~\cite{gu-etal-2024-middleware,xin-etal-2025-atomr,xiong-etal-2025-multi}.

Existing planning techniques can be categorized into two major paradigms by \textbf{planning horizon}, the number of steps planned before tool execution~(Figure~\ref{fig:introduction-pne-vs-itr}). \textit{Single-step horizon (SH)} planning interleaves reasoning and execution, calling one tool at a time based on prior observations. \textit{Full-horizon (FH)} planning instead generates a complete plan upfront before execution. Recent frontier models and systems increasingly support upfront high-level task decomposition, and more advanced FH planning techniques have also been developed~\cite{xu-etal-2023-rewoo,li-etal-2025-agent}. However, tight ``think-act-observe'' loops~\cite{yao-etal-2023-react} remain a common default for low-level tool execution.\footnote{In this paper, we focus on this \textbf{low-level execution layer} rather than higher-level decomposition.} This \textit{eager} monitoring is often presumed essential for robustly handling the opacity and potential noise of external tools and data sources~\cite{kim-et-al-2024-llm,gonzalez-pumariega-etal-2025-robotouille,zhang-etal-2025-recap}.

Recent work by \citet{liu-etal-2025-select-decompose} has begun to question whether interleaved planning is universally optimal. They compare different planning strategies on general reasoning tasks without tool interaction and find that SH planning is not consistently superior. However, their study leaves open a critical question: does this conclusion extend to data-centric settings, where success depends on structured tool calls rather than purely internal reasoning?

We perform a controlled empirical study and address this gap by shifting the focus from abstract reasoning to \textbf{data-centric, tool-calling tasks}, in which planning decisions directly affect execution success, computational cost, and robustness. We further go beyond dataset-level comparisons by analyzing planning behavior at the instance level, enabling a more precise characterization of task difficulty. Specifically, we introduce an instance-level framework that disentangles two orthogonal dimensions: \textbf{topological complexity} (the depth and breadth of the execution graph) and \textbf{tool robustness} (the tolerance to imprecise inputs). Together, these dimensions capture two fundamental challenges in tool-mediated data-centric tasks: satisfying logical dependencies among intermediate steps and aligning generated arguments with external schema or vocabulary constraints.

We hypothesize that for well-defined data-centric tasks, an FH planner equipped with \textit{lazy} monitoring (executing a complete plan and replanning only upon failure) can match the performance of an SH planner without the massive overhead of continuous feedback integration. We focus on Knowledge Base Question Answering (KBQA) and Multi-hop QA (HotpotQA) because they represent the two core challenges of data-centric agents. KBQA serves as a controlled environment for testing logic-alignment, requiring agents to coordinate atomic tool operations that mirror database queries. HotpotQA represents the challenges in unstructured settings, where agents must navigate retrieval noise and coordinate reasoning-based sub-agents. This setup allows us to analyze planning behavior across both rigid, structured schemas and fuzzier, unstructured data sources.

Our results show no statistical evidence that SH planning provides a performance advantage over FH planning across varying structural or robustness configurations in well-defined data-centric tasks. Given this performance parity, the $2$--$3\times$ efficiency advantage of FH planning~(Figure~\ref{fig:introduction-comparison}) makes it a stronger option than previously thought. Our findings also suggest that the perceived brittleness of FH planning in prior work may result from the absence of proper recovery mechanisms rather than an inherent limitation of full-horizon planning. While SH planning may remain advantageous in exploratory or highly dynamic tool-calling tasks, our results refine prevailing assumptions about SH by demonstrating that eager monitoring is not universally necessary. In structured and stable environments, less frequent monitoring can substantially improve efficiency without sacrificing accuracy.

Our contributions are the following:
\begin{itemize}
\item We isolate planning horizon as a core architectural variable in LLM agents~(\S\ref{sec:method-characterization}) and provide a controlled comparison of its effects in data-centric tasks.
\item We introduce an instance-level framework that characterizes difficulty via execution graph topology and tool robustness~(\S\ref{sec:task-characterization}). In particular, we identify depth and breadth as overlooked axes of execution-graph complexity that affect planning performance beyond sequential length.
\item We show that FH planning with lazy replanning achieves accuracy parity with SH planning while using fewer tokens on well-defined data-centric tasks~(\S\ref{sec:experiments}). This result provides a foundation for future work on adaptive and hybrid planners.
\end{itemize}

\section{Planning Horizon : The Key Architectural Feature}\label{sec:method-characterization}

The landscape of agentic frameworks is expansive~\cite{huang-etal-2024-understanding,li-2025-review,wei-etal-2025-plangenllms}. Yet, most approaches can be understood through a single underlying design choice: \textbf{planning horizon}, defined as the number of steps an agent plans before execution. Planning horizon determines how much simulation an agent performs prior to interacting with tools and therefore governs when feedback from the environment is incorporated into planning.

We conceptualize the agent as a policy $\pi$ interacting with a tool execution environment $\mathbb{E}$. Let $\mathbb{A}$ denote the set of available tools (actions) and $\mathbb{O}$ the space of possible observations (tool outputs). Given a user query $q$, the agent produces a plan $\mathcal{P} = (a_1, \dots, a_T)$ consisting of actions $a_i \in \mathbb{A} (i = 1,\dots,T)$, where $T$ denotes the plan length and $a_T$ is the final planned action. Here, a plan refers specifically to a sequence of tool calls at the execution level. Executing an action $a_i$ yields an observation $o_i \in \mathbb{O}$, resulting in a trajectory that leads to the answer to $q$. Within this formulation, planning strategies differ only in when the policy $\pi$ is invoked during execution.

In this section, we treat planning horizon as a primary architectural feature and discuss how it shapes the trade-off between adaptivity and cost. In particular, we focus on when and how agents replan in response to tool feedback.

\subsection{Single-Step Horizon (SH)}

The single-step horizon (SH) paradigm tightly interleaves planning and execution. The agent simulates only one step ahead before performing an action. In this paradigm, action generation at step $t$ is conditioned on the user query $q$ and the history of previous actions and observations:
\begin{align}
    a_t \sim \pi(q, a_1, o_1, ..., a_{t-1}, o_{t-1}) \label{eq:sh-planning}
\end{align}
This design entails \textit{eager} feedback monitoring, since the agent \textit{always} processes observations up to the current step before deciding on the next action.

SH is widely used in modern agentic applications, with some variations that can plan concurrent, independent steps at once. Its popularity stems from the tight planning--acting feedback loop, which enables robust adaptation to uncertainty and noise in external tools and environments. Such adaptivity is particularly beneficial in exploratory tasks that require information gathering before subsequent planning~\cite{kim-et-al-2024-llm,gonzalez-pumariega-etal-2025-robotouille,zhang-etal-2025-recap}.

\subsection{Full-Horizon (FH)}

The full-horizon (FH) paradigm generates a complete execution graph upfront. The agent performs a full simulation of the trajectory required to solve the task before it triggers any tool execution. In this paradigm, the agent generates an initial plan as a complete sequence of actions $\mathcal{P} = (a_1, \dots, a_T)$ conditioned only on the query:
\begin{align}
    \mathcal{P} \sim \pi(q) \label{eq:fh-planning}
\end{align}

Execution then proceeds via the environment $\mathbb{E}$ without invoking the policy $\pi$ at every step. Unlike SH, FH can allow \textit{lazy} feedback integration, wherein observations are incorporated only when monitoring is triggered. For example, an execution failure at step $k$ can trigger monitoring, prompting the agent to replan based on the observed trajectory:
\begin{align}
    \mathcal{P}' \sim \pi(q, a_1, o_1, ..., a_{k}, o_{k}) \label{eq:fh-replanning}
\end{align}

\subsection{Implications of Planning Horizon}

Planning horizon creates a trade-off between adaptivity and computational cost. SH \textit{eagerly} monitors every tool call, which can yield robust adaptation to uncertainty and noise but incurs substantial overhead due to repeated inference. FH, by contrast, generates a multi-step plan in a single pass and is therefore more efficient, but only integrates feedback \textit{lazily}.

Comparing FH’s lazy feedback integration (Eq.~\eqref{eq:fh-replanning}) with SH’s single-step generation (Eq.~\eqref{eq:sh-planning}), we see that they both condition on the same action--observation history available at that step. Thus, the practical difference is not \emph{what} information is used, but \emph{when} it is used: at every step (SH) versus only when triggered (FH).

This insight suggests a testable hypothesis: with an appropriate monitoring trigger, FH should achieve performance comparable to SH while requiring fewer policy calls. From this perspective, the commonly assumed brittleness of FH may arise not from its planning horizon per se, but from the absence of effective error recovery mechanisms. Notably, many existing studies implement FH without replanning~\cite{gonzalez-pumariega-etal-2025-robotouille,zhang-etal-2025-recap}, which confounds evaluation of the paradigm.

We therefore evaluate FH equipped with lazy monitoring. To obtain generalizable insights, we adopt a simple trigger that is commonly used in practice: replanning upon execution failure. We also abstract away implementation-specific details such as tool-calling formats and isolate planning horizon as the primary variable in our analysis.

\section{Task Characterization}\label{sec:task-characterization}

To evaluate how planning horizon shapes agent behavior, we require a principled way to characterize task difficulty. Dataset-level comparison~\cite{liu-etal-2025-select-decompose} can obscure meaningful variation across instances. Instances actually differ substantially in which tools and data sources are involved and how they are connected. Without explicitly modeling such \textbf{instance-level} features, comparisons between planning strategies may conflate planning effects with underlying instance properties.

In data-centric tool-calling tasks, success depends on \emph{aligning logic and vocabulary}: correctly composing tool calls to match the latent dependency structure of the task instance (\textit{logic alignment}) and specifying tool parameters that match the schema or representation of the external data source (\textit{vocabulary alignment}). capture these challenges using two independent dimensions: topological complexity and tool robustness. This separation allows us to disentangle structural reasoning difficulty from environmental uncertainty. These factors are often conflated in prior evaluations.

\subsection{Topological Complexity (Logic Alignment)}\label{sec:topological-complexity}

\begin{figure}[t]
    \centering
    \includegraphics[width=1.0\linewidth]{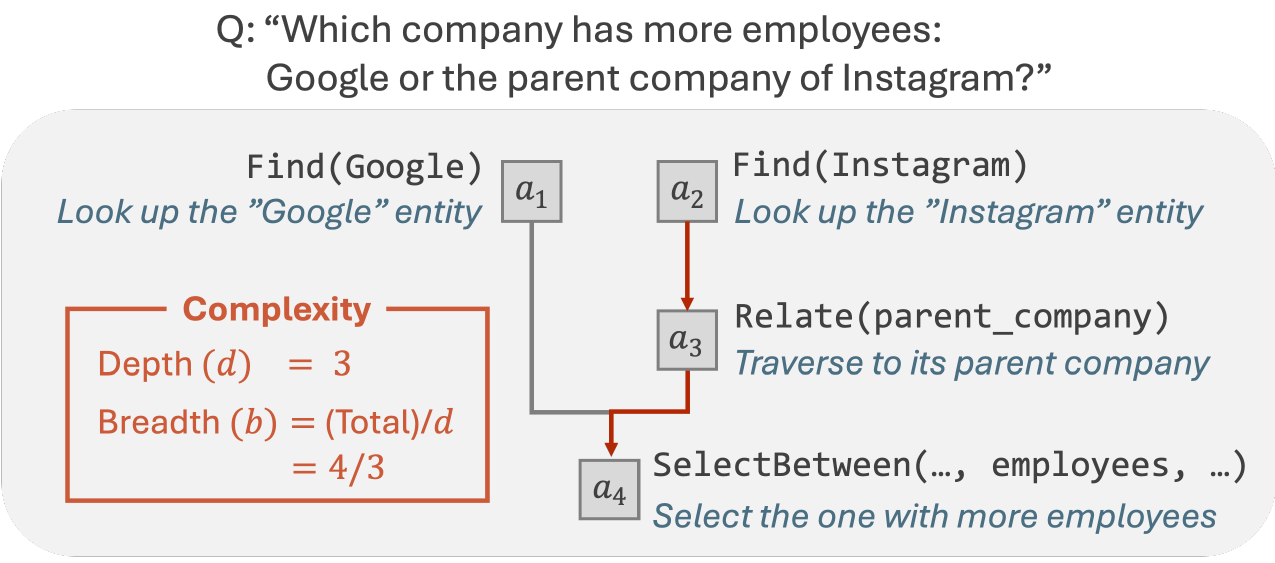}
    \caption{An example plan DAG for the query: ``Which company has more employees: Google or the parent company of Instagram?''. The DAG involves two parallel paths: a single-hop lookup for Google ($a_1$) and a two-hop traversal for Instagram's parent company ($a_2 \rightarrow a_3$), which are finally compared at the aggregation node ($a_4$). This structure results in a depth of $d=3$ and a breadth of $b=4/3$.}
    \label{fig:topological-complexity}
    \Description{A directed acyclic graph illustrating the execution plan for comparing the employee counts of Google and Instagram's parent company. One branch performs a single-hop lookup for Google, while the other performs a lookup followed by a parent-company traversal for Instagram. The two branches are merged by a final comparison node. The figure highlights a critical-path depth of three steps and an average breadth of four-thirds.}
\end{figure}

Topological complexity captures the difficulty of aligning execution with the latent logical structure of a task. We represent a plan as a directed acyclic graph (DAG) $G = (V, E)$, where nodes ($V$) are tool calls and edges ($E$) are dependencies between them. Motivated by task scheduling studies~\cite{sevcik-1989-characterization,calzarossa-and-serazzi-1993-workload}, we characterize each execution graph using two metrics:
\begin{description}
    \item[Depth ($d$):] Depth is defined as the critical path length of the DAG, i.e., the longest chain of dependent tool calls. Depth captures the minimum number of sequential reasoning steps that must be executed in order. High depth increases exposure to cascading errors.
    \item[Breadth ($b$):] We define breadth as the average parallelism of the graph: $b = |V|/d$. Higher breadth indicates the presence of multiple independent sub-plans that must eventually be merged. LLM planners typically operate over a linearized, text-form plan. This technical constraint can make it difficult to manage highly branched plans.
\end{description}

Consider the query: ``Which company has more employees: Google or the parent company of Instagram?'' Figure~\ref{fig:topological-complexity} illustrates the plan DAG for this task using the KoPL tools (Appendix~\ref{app:task-details}). The query involves two parallel branches (a single entity lookup and an entity lookup followed by a relationship traversal). These branches are then merged at an aggregation node to perform the final comparison. The critical path is $a_2 \rightarrow a_3 \rightarrow a_4$, yielding a depth of $d=3$. With $|V|=4$ total steps, the parallel breadth is $b=4/3$. This example indicates that structural complexity arises not only from sequential length but also from the need to coordinate multiple independent sub-queries.

\subsection{Tool Robustness (Vocabulary Alignment)}\label{sec:tool-robustness}

Even when a plan's logical structure is correct, execution can fail if tool inputs do not exactly match the representation of data sources. We thus define tool robustness to measure how tolerant an environment is to imperfect tool specifications. We consider two common forms of robustness:
\begin{description}
    \item[Robustness to Schema Mismatch:] In structured data environments (e.g., KBQA), tool parameters must match predefined schema elements such as relation names or attribute keys. For instance, a tool call that uses \texttt{employees} may fail if the schema only contains \texttt{employee\_counts}. This mismatch can be mitigated when the tool supports flexible recovery mechanisms such as embedding-based soft matching (e.g., mapping \texttt{employees} to \texttt{employee\_counts} given the query context).
    \item[Robustness to Retrieval Uncertainty:] In unstructured environments (e.g., document retrieval), failure may arise from search mismatch rather than schema mismatch. %
    For example, the query ``who \textit{owns} Instagram'' may not retrieve evidence that states ``Instagram \textit{was acquired} by Meta'', depending on the retrieval quality. Robustness can vary depending on recall of the retriever (top-1 vs top-100), semantic matching quality and evidence aggregation strategy. %

\end{description}

Although other robustness factors can also be meaningful, such as tolerance to technical failures (e.g., service downtime), we focus on the semantic and informational dimensions above because they are particularly common in data-centric QA tasks.

\section{Experiments}\label{sec:experiments}

We aim to assess whether step-wise monitoring is necessary for complex data-centric tool use, or whether comparable performance is achievable with FH with lazy monitoring. For this purpose, we compare SH and FH planning with lazy under controlled variation in topological complexity and tool robustness. Specifically, we address three questions: (1) Does SH planning outperform FH planning in overall accuracy? (2) Does SH planning better handle greater topological complexity (deep/wide graphs)? (3) Does SH planning exhibit greater robustness in noisy environments?

\subsection{Experimental Setup}

\begin{figure}[t]
    \centering
    \includegraphics[width=1\linewidth]{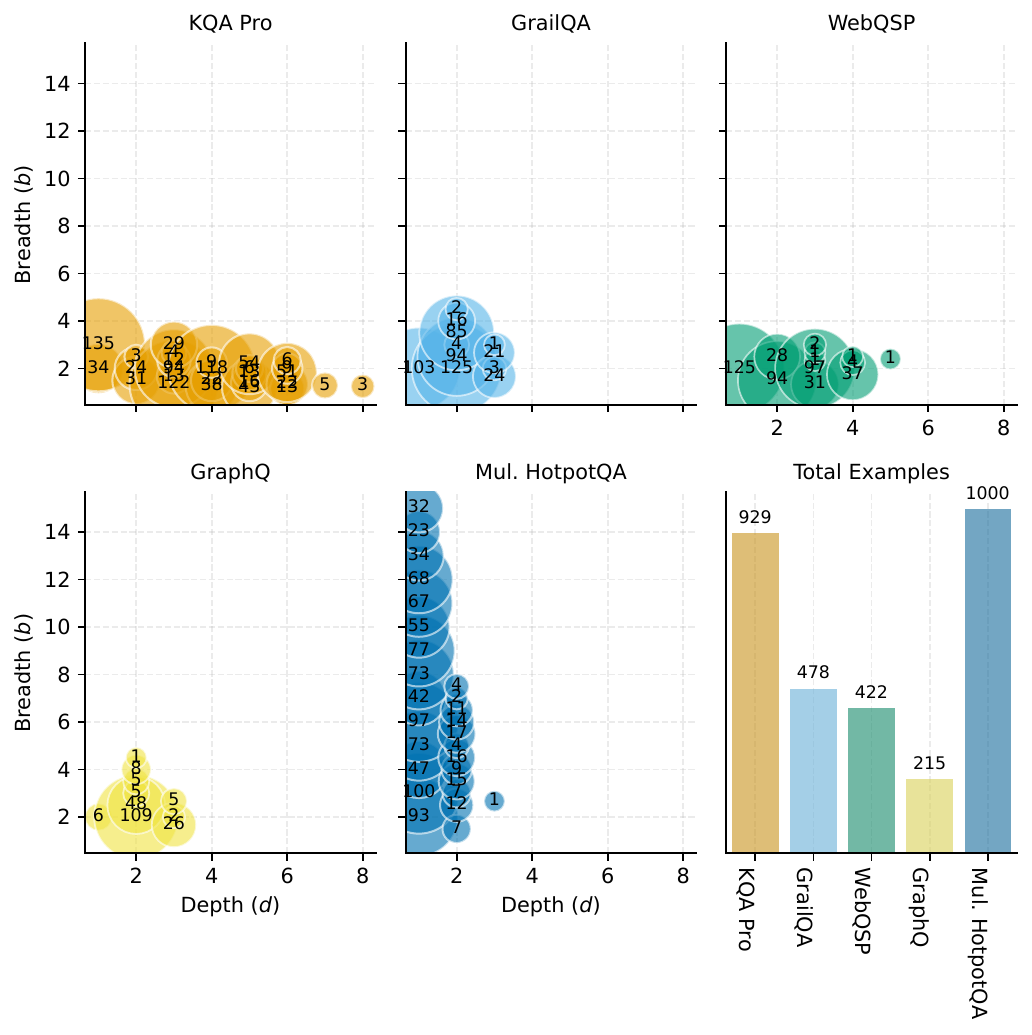}
    \caption{Distribution of task instances across datasets. The horizontal and vertical axes represent the critical path length (depth) and average parallelism (breadth), respectively. The bar chart summarizes total instance counts per dataset.}
    \label{fig:dataset-spec}
    \Description{A dataset overview figure combining a scatter plot and a bar chart. The scatter plot places task instances by execution-graph depth on the horizontal axis and average parallelism breadth on the vertical axis, showing how instances from different datasets are distributed across topological complexity. The accompanying bar chart summarizes the total number of instances in each dataset.}
\end{figure}

\begin{table}[t]
\centering

\caption{Example task instances with their corresponding ground-truth tool-call trajectories (execution graphs).}
\label{tab:task-examples}
\begin{minipage}{0.9\linewidth}
    \subcaption{KQA Pro}
    \label{tab:task-example-kqapro}
    \begin{tabular}{p{1\textwidth}}
    \toprule
    Q: Who is taller, LeBron James Jr. or his father? \\
    \midrule
    Solution (Simplified) \\
    \begin{minipage}{1\linewidth}
        \centering
        \vspace{0.25\baselineskip}
        \includegraphics[height=0.85in]{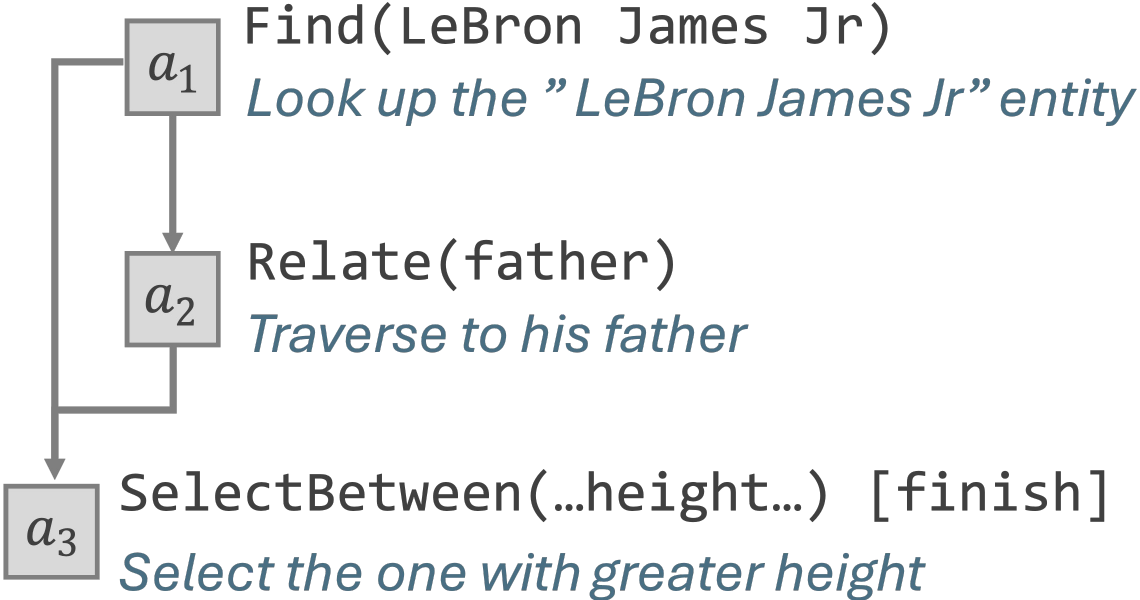}
        \vspace{0.25\baselineskip}
    \end{minipage}\\
    A: LeBron James Jr. \\
    \bottomrule
    \end{tabular}
\end{minipage}\\ \vspace{1\baselineskip}
\begin{minipage}{0.9\linewidth}
    \subcaption{GrailQA}
    \label{tab:task-example-grailqa}
    \begin{tabular}{p{1\textwidth}}
    \toprule
    Q: What movie with a television running time of less than 60 minutes features Taylor Lautner? \\
    \midrule
    Solution (Simplified) \\
    \begin{minipage}{1\linewidth}
        \centering
        \vspace{0.25\baselineskip}
        \includegraphics[height=0.85in]{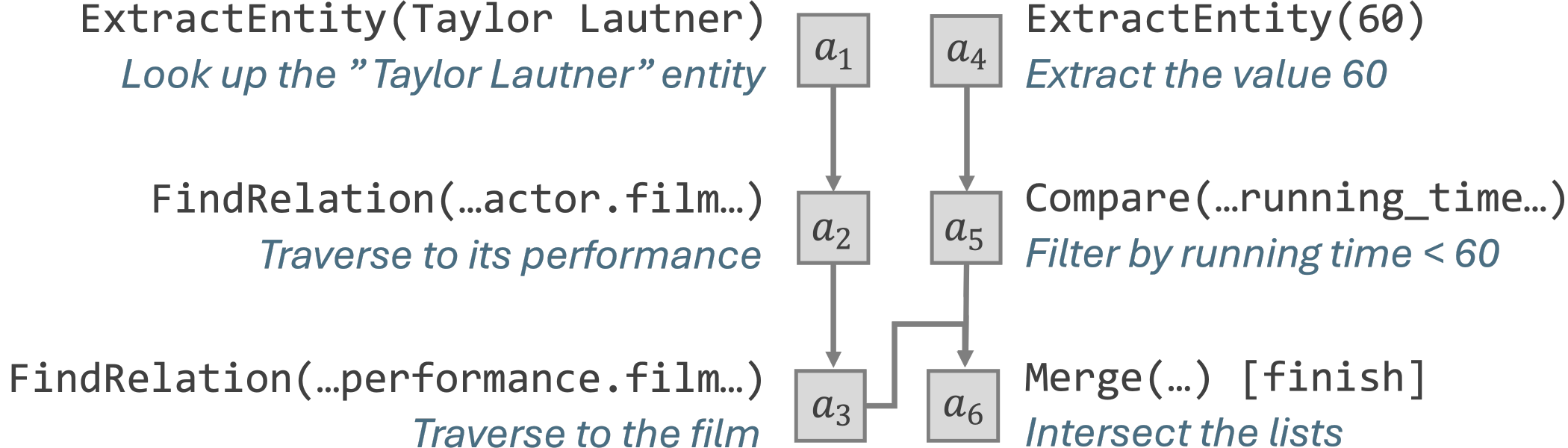}
        \vspace{0.25\baselineskip}
    \end{minipage}\\
    A: m.02686wj (``He's a Bully, Charlie Brown'') \\
    \bottomrule
    \end{tabular}
\end{minipage}\\ \vspace{1\baselineskip}
\begin{minipage}{0.9\linewidth}
    \subcaption{Multi-objective HotpotQA ($k=2$)}
    \label{tab:task-example-hotpotqa}
    \begin{tabular}{p{1\textwidth}}
    \toprule
    Q: 1. Eugeniusz Bodo and Chris Buck both shared what occupation? 2. The creator of the record label Merciful Release was born in what year? \\
    \midrule
    Solution (Simplified) \\
    \begin{minipage}{1\linewidth}
        \centering
        \vspace{0.25\baselineskip}
        \includegraphics[height=0.85in]{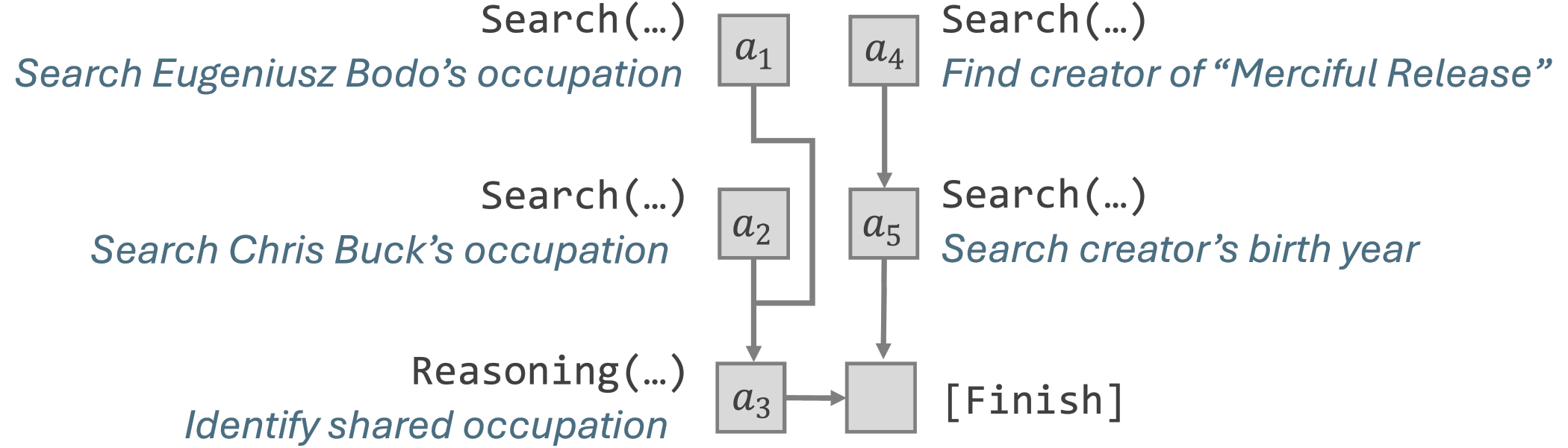}
        \vspace{0.25\baselineskip}
    \end{minipage}\\
    A: 1. film director, 2. 1959 \\
    \bottomrule
    \end{tabular}
\end{minipage}

\end{table}

We evaluate performance across data-centric tasks over two types of underlying data:

\begin{description}
\item[KBQA (Structured):] We use KQA Pro~\cite{cao-etal-2022-kqa}, GrailQA~\cite{gu-etal-2021-beyond}, WebQSP~\cite{yih-etal-2016-value}, and GraphQ~\cite{su-etal-2016-generating} as our main testbed. These tasks require precise logical composition over structured knowledge bases (Wikidata and Freebase), where answers are constructed through multi-step queries. For KQA Pro, we employ KoPL tools (27 tools). For the rest, we use the atomic query tools~\cite{luo-etal-2025-kbqa} (7 tools). A tool execution is considered as failed if it returns no valid result.
    \item [Multi-objective HotpotQA (Unstructured):] 
    Following \citet{zhou-etal-2025-mem1}, we combine examples from HotpotQA~\cite{yang-etal-2018-hotpotqa} to synthesize multi-objective questions with greater breadth ($k \in \{2, 3, 4, 5\}$), and also include the original HotpotQA instances ($k=1$). This setting requires accurate task decomposition and search over Wikipedia. We use the search and reasoning sub-agents from \citet{kim-etal-2024-husky} as tools. Each sub-agent takes a short sub-question as input and returns a short answer if supporting evidence is retrieved (search) or if the reasoning is successfully performed (reasoning). Note that modern LLMs can answer many HotpotQA questions without tools. We include this dataset primarily to connect our analysis with prior work.
\end{description}

To enable analysis based on topological features, we derive ground-truth execution trajectories from dataset provenance. For KBQA, we use the gold solution program (KQA Pro) or logical forms (GrailQA, WebQSP, and GraphQ) to obtain the sequence of tool calls with dependency information. For HotpotQA, we use gold supporting documents to determine the required retrieval and reasoning steps, assisted by LLM annotation (GPT-4.1-mini). We downsample all datasets to balance reasoning length and question types, yielding 929 KQA Pro instances, 1,115 Atomic KBQA instances, and 1,000 multi-objective HotpotQA instances.

As discussed in Section~\ref{sec:tool-robustness}, we model tool robustness using schema matching in KBQA and the retrieval rank cutoff (top-(k)) in multi-objective HotpotQA. For the main results~(\S\ref{sec:exp-main}) and topological complexity analysis~(\S\ref{sec:exp-topological-complexity}), we use high-robustness settings, where tools are more tolerant to imprecise input parameters. In KBQA, when a parameter specified by the planner does not exactly match the KB schema, we perform soft schema matching by (i) retrieving candidates via cosine similarity over BGE embeddings~\cite{xiao-etal-2024-c-pack} and (ii) validating candidates with GPT-4.1-mini. If no valid match is found, the tool returns the top-10 nearest candidates as feedback. In multi-objective HotpotQA, the search tool retrieves the top-10 documents to increase the chance of retrieving the necessary evidence. In controlled robustness experiments~(\S\ref{sec:exp-tool-robustness}), we switch to low-robustness settings by disabling soft matching in KBQA or restricting retrieval to top-1 in HotpotQA.

Figure~\ref{fig:dataset-spec} shows the distribution of task instances across datasets, and Table~\ref{tab:task-examples} provides examples with ground-truth trajectories. More details on dataset-specific pre-processing and tool implementations can be found in Appendix~\ref{app:task-details}.

\paragraph{Backbone LLMs:} We compare four backbone LLMs representing distinct capability classes: (1) GPT-4.1-mini and Qwen3-235B-A22B (Instruct), representing standard instruction-tuned models; and (2) GPT-5-mini and Gemini-3-Flash, representing frontier reasoning models with stronger inherent planning capabilities.

\paragraph{Metrics:} We repeat the same configuration three times to account for the randomness of LLMs. We report average accuracy (using LLM-as-judge for soft matching) and efficiency (average input/output tokens per query). See Appendix \ref{app:evaluation-details} for details.

\begin{lstlisting}[style=booktabsstyle, caption={Planning prompts for Atomic KBQA}, label={prmt:atomic-kbqa}]
[system] (FH)
Use the available tools to answer the user's question step by step, generating the entire plan. Use $i to refer to the output of step i (0-indexed).

**Tools**
{tool_definitions}

You can access the knowledge base through the provided tools.

**Examples**
{demonstrations}

[system] (SH)
Use the available tools to answer the user's question step by step, generating one action at a time. Use $i to refer to the output of step i (0-indexed).

(...same as FH)

[user]
Question: {question}
\end{lstlisting}

\paragraph{Implementation Details:} As discussed in Section~\ref{sec:method-characterization}, FH and SH differ primarily in planning horizon. To ensure a fair comparison, we use minimal prompts that include task instructions, tool definitions (as a JSON string), and in-context demonstrations (see Prompt~\ref{prmt:atomic-kbqa}), while keeping all other components identical across paradigms. For each task instance, the agent is allowed to execute up to 30 tool calls, and if tool execution fails, the agent can retry up to eight times. We constrain the LLM decoding to produce a plan as a sequence of tool calls in JSON format. We do not use native tool-calling APIs, which are not applicable to FH. For in-context learning, we sample 10 demonstration examples from the training set and provide the same set to both paradigms, retrieving semantically similar examples via query embedding. Details such as LLM prompts and hyper-parameters can be found in Appendix~\ref{app:planner-details}.

\subsection{Overall Accuracy and Efficiency}\label{sec:exp-main}

\begin{table*}[tb]
\centering

\caption{Comparison of FH and SH across structured (KBQA) and unstructured (multi-objective HotpotQA) tool-calling tasks. FH matches SH in accuracy across most settings while using substantially fewer input tokens than SH. The efficiency gap widens as the tool set grows (e.g. KQA Pro).}
\label{tab:main-results}

{\renewcommand{\arraystretch}{1.2}
\begin{tabular}{llrrrrrrrcc}
\toprule
 &  & \multicolumn{3}{c}{Accuracy ($\uparrow$)} & \multicolumn{2}{c}{Input Token ($\downarrow$)} & \multicolumn{2}{c}{Output Token ($\downarrow$)} & \multicolumn{2}{c}{Ratio (SH/FH)} \\
\cmidrule{3-11}
Dataset & Model & \multicolumn{1}{c}{FH} & \multicolumn{1}{c}{SH} & \multicolumn{1}{c}{$\Delta$SH} & \multicolumn{1}{c}{FH} & \multicolumn{1}{c}{SH} & \multicolumn{1}{c}{FH} & \multicolumn{1}{c}{SH} & \multicolumn{1}{c}{Input} & \multicolumn{1}{c}{Output} \\
\midrule
\rowcolor[gray]{0.95}

\multirow[t]{4}{*}{KQA Pro} & GPT-4.1-mini & 0.804 & 0.806 & \poscolor{+0.002} & 43,719 & 119,431 & 346 & 198 & 2.7 & 0.6 \\
\rowcolor[gray]{0.95}
 & GPT-5-mini & 0.901 & 0.912 & \poscolor{+0.011} & 21,144 & 97,239 & 1,725 & 3,577 & 4.6 & 2.1 \\
\rowcolor[gray]{0.95}
 & Qwen3-235B-A22B & 0.834 & 0.848 & \poscolor{+0.014} & 18,140 & 67,008 & 417 & 234 & 3.7 & 0.6 \\
\rowcolor[gray]{0.95}
 & Gemini-3-Flash & 0.908 & 0.870 & \negcolor{-0.038} & 16409 & 77867 & 542 & 339 & 4.7 & 0.6 \\

\multirow[t]{4}{*}{GrailQA} & GPT-4.1-mini & 0.725 & 0.726 & \poscolor{+0.001} & 14,534 & 29,142 & 331 & 166 & 2.0 & 0.5 \\
 & GPT-5-mini & 0.804 & 0.782 & \negcolor{-0.023} & 11,224 & 25,665 & 2,424 & 3,606 & 2.3 & 1.5 \\
 & Qwen3-235B-A22B & 0.722 & 0.740 & \poscolor{+0.017} & 16,143 & 25,849 & 541 & 240 & 1.6 & 0.4 \\
 & Gemini-3-Flash & 0.845 & 0.691 & \negcolor{-0.154} & 14,493 & 51,923 & 804 & 550 & 3.6 & 0.7 \\
 
\rowcolor[gray]{0.95}
\multirow[t]{4}{*}{WebQSP} & GPT-4.1-mini & 0.751 & 0.729 & \negcolor{-0.022} & 7,092 & 22,391 & 195 & 127 & 3.2 & 0.7 \\
\rowcolor[gray]{0.95}
 & GPT-5-mini & 0.771 & 0.759 & \negcolor{-0.012} & 5,887 & 21,781 & 1,004 & 2,071 & 3.7 & 2.1 \\
\rowcolor[gray]{0.95}
 & Qwen3-235B-A22B & 0.780 & 0.784 & \poscolor{+0.004} & 6,098 & 18,809 & 301 & 181 & 3.1 & 0.6 \\
\rowcolor[gray]{0.95}
 & Gemini-3-Flash & 0.815 & 0.799 & \negcolor{-0.016} & 5,732 & 20,657 & 427 & 283 & 3.6 & 0.7 \\

\multirow[t]{4}{*}{GraphQ} & GPT-4.1-mini & 0.535 & 0.540 & \poscolor{+0.005} & 20,535 & 32,665 & 445 & 193 & 1.6 & 0.4 \\
 & GPT-5-mini & 0.704 & 0.702 & \negcolor{-0.002} & 13,725 & 27,189 & 3,138 & 4,537 & 2.0 & 1.4 \\
 & Qwen3-235B-A22B & 0.620 & 0.594 & \negcolor{-0.026} & 20,643 & 27,551 & 627 & 258 & 1.3 & 0.4 \\
 & Gemini-3-Flash & 0.749 & 0.577 & \negcolor{-0.172} & 16,021 & 55,958 & 840 & 585 & 3.5 & 0.7 \\

\rowcolor[gray]{0.95}
\multirow[t]{4}{*}{Mul. HotpotQA} & GPT-4.1-mini & 0.572 & 0.604 & \poscolor{+0.032} & 22,017 & 31,737 & 1,099 & 223 & 1.4 & 0.2 \\
\rowcolor[gray]{0.95}
 & GPT-5-mini & 0.605 & 0.593 & \negcolor{-0.012} & 29,062 & 51,442 & 4,526 & 10,086 & 1.8 & 2.2 \\
\rowcolor[gray]{0.95}
 & Qwen3-235B-A22B & 0.517 & 0.615 & \poscolor{+0.098} & 23,911 & 33,917 & 1,390 & 305 & 1.4 & 0.2 \\
\rowcolor[gray]{0.95}
 & Gemini-3-Flash & 0.545 & 0.519 & \negcolor{-0.026} & 29,348 & 57,118 & 1,844 & 622 & 1.9 & 0.3 \\ 
\bottomrule
\end{tabular}}
\end{table*}

Table~\ref{tab:main-results} presents aggregate accuracy and token consumption for both paradigms across all datasets and LLM backends.

\paragraph{Accuracy:} SH shows no clear performance advantage over FH. In fact, FH significantly outperforms SH on Atomic KBQA datasets for Gemini-3-Flash (e.g., SH is worse by 15.4 accuracy points on GrailQA and 17.2 points on GraphQ). We provide a more detailed analysis in Section~\ref{sec:exp-failure-mode}. This pattern result suggests that, for this specific reasoning model, FH planning can be more effective than SH planning.

\paragraph{Efficiency:} FH consistently reduces input token consumption by 2--3x. The magnitude of this reduction depends on the length of tool definitions in the input prompt. For KQA Pro (27 tools with complex schemas), the savings are massive (2.7--4.7x). For HotpotQA (2 tools), the margin is smaller (1.4--1.9x) but still significant. Crucially, for reasoning models (GPT-5-mini and Gemini-3-Flash), FH also saves output tokens. While SH triggers a full Chain-of-Thought generation at every single step, FH generates the reasoning chain only once during initial plan generation and during occasional replanning. This efficiency advantage is particularly beneficial in production settings, where token usage directly translates to latency and cost.

\paragraph{Does SH outperform FH in overall accuracy?} No. Overall, SH does not deliver a clear accuracy advantage over FH, while FH is more token-efficient. We next analyze performance under controlled variation in topological complexity and tool robustness.

\subsection{Topological Complexity Analysis}\label{sec:exp-topological-complexity}

We employ a logistic regression model to quantify the impact of planning strategy while controlling for task complexity. Because we have repeated measures (three runs per query), we fit the model with Generalized Estimating Equations (GEE) using clustering by question ID. The specification is:
\begin{align}
    \text{logit}(P(y=1)) = & \beta_0 + \beta_d d^* + \beta_b b^* + \beta_\text{SH} x_\text{SH} \nonumber \\
    & + \beta_{d:\text{SH}} (d^* \times x_\text{SH}) + \beta_{b:\text{SH}} (b^* \times x_\text{SH}) + \dots_{,} \nonumber
\end{align}
where $y \in \{0, 1\}$ indicates task success, $\beta$ denotes coefficients to be estimated, $d^*$ and $b^*$ are standardized depth and breadth features~(\S\ref{sec:topological-complexity}), and $x_\text{SH} \in \{0, 1\}$ is a binary indicator for the SH strategy. We additionally include dataset-specific control variables, for example the final reasoning operation type (e.g., \texttt{VerifyDate}) for KBQA, dataset identity for Atomic KBQA, and question type (bridge/comparison) for HotpotQA. For brevity, we omit these controls from the summary tables in this section. Full model specifications and results are provided in Appendix~\ref{app:evaluation-details} and Appendix~\ref{app:gee-results}.

\begin{table*}[tb]
\centering

\caption{Summary of GEE coefficients ($**p<0.01$, $*p<0.05$). Atomic KBQA results include GrailQA, WebQSP, and GraphQ.}
\label{tab:glm-coefficients}

\begin{tabular}{llllllll}
\toprule
 & & & & & \multicolumn{2}{c}{Interaction} \\
\cmidrule{6-7}
Dataset & Model & Depth ($\beta_d$) & Breadth ($\beta_b$) & Is SH? ($\beta_\text{SH}$) & $\beta_{d:\text{SH}}$ & $\beta_{b:\text{SH}}$ \\
\midrule

\rowcolor[gray]{0.95}
\multirow[t]{4}{*}{KQA Pro} & GPT-4.1-mini & $-0.646^{**}$ & $-0.423^{**}$ & $+0.081$ & $-0.238^{**}$ & $-0.283^{**}$ \\
\rowcolor[gray]{0.95}
 & GPT-5-mini & $-0.496^{**}$ & $-0.464^{*}$ & $+0.132$ & $+0.047$ & $+0.091$ \\
\rowcolor[gray]{0.95}
 & Qwen3-235B-A22B & $-0.685^{**}$ & $-0.426^{**}$ & $+0.178^{*}$ & $+0.011$ & $+0.301^{**}$ \\
\rowcolor[gray]{0.95}
 & Gemini-3-Flash & $-0.396^{*}$ & $-0.465^{*}$ & $-0.413^{**}$ & $-0.111$ & $+0.067$ \\

\multirow[t]{4}{*}{Atomic KBQA} & GPT-4.1-mini & $-0.247^{**}$ & $-0.159^{*}$ & $-0.017$ & $-0.075$ & $-0.076$ \\
 & GPT-5-mini & $-0.261^{**}$ & $-0.226^{**}$ & $-0.084^{*}$ & $+0.023$ & $+0.017$ \\
 & Qwen3-235B-A22B & $-0.243^{**}$ & $-0.314^{**}$ & $+0.019$ & $-0.012$ & $+0.036$ \\
 & Gemini-3-Flash & $-0.261^{**}$ & $-0.139$ & $-0.589^{**}$ & $-0.005$ & $-0.196^{**}$ \\

\rowcolor[gray]{0.95}
\multirow[t]{4}{*}{Mul. HotpotQA} & GPT-4.1-mini & $-0.256^{**}$ & $-0.576^{**}$ & $+0.152^{**}$ & $-0.002$ & $-0.051$ \\
\rowcolor[gray]{0.95}
 & GPT-5-mini & $-0.275^{**}$ & $-0.470^{**}$ & $-0.057$ & $+0.034$ & $+0.033$ \\
\rowcolor[gray]{0.95}
 & Qwen3-235B-A22B & $-0.158^{**}$ & $-0.426^{**}$ & $+0.458^{**}$ & $-0.070$ & $-0.124^{*}$ \\
\rowcolor[gray]{0.95}
 & Gemini-3-Flash & $-0.179^{**}$ & $-0.334^{**}$ & $-0.103$ & $-0.142^{*}$ & $-0.245^{**}$ \\
\bottomrule
\end{tabular}
\end{table*}

\paragraph{Result:} As shown in Table~\ref{tab:glm-coefficients}, the coefficients for depth ($\beta_{d}$) and breadth ($\beta_{b}$) are consistently negative and statistically significant ($p < 0.01$) across most datasets and models. This result confirms that topological complexity is a challenging factor that degrades performance \textit{regardless of the planning paradigm.}

\paragraph{Does SH handle greater topological complexity (deep/wide graphs) better than FH?} In general, no. The interaction coefficients between planning method and topological complexity ($\beta_{d:\text{SH}}$ and $\beta_{b:\text{SH}}$) are statistically insignificant ($p > 0.05$) in most settings. In some cases, SH is even \textit{worse} at handling increasing complexity (e.g., GPT-4.1-mini on KQA Pro). Nevertheless, these differences are small compared to the dominant negative impact of topological complexity ($\beta_d$ and $\beta_b$) overall.

\subsection{Tool Robustness Analysis}\label{sec:exp-tool-robustness}

\begin{table}[t]
\centering

\caption{Performance change (absolute accuracy difference) under low robustness settings. Difference larger than 0.05 is denoted in \textbf{bold.}}
\label{tab:tool-robustness}

\begin{tabular}{l@{\hspace{4pt}}lrr}
\toprule
 & & \multicolumn{2}{c}{$\Delta$Accuracy} \\
 \cmidrule{3-4}
Dataset & Model & \multicolumn{1}{c}{FH} & \multicolumn{1}{c}{SH} \\
\midrule
\rowcolor[gray]{0.95}
\multirow[t]{4}{*}{KQA Pro} & GPT-4.1-mini & 0.000 & -0.014 \\
\rowcolor[gray]{0.95}
 & GPT-5-mini & 0.012 & 0.003 \\
\rowcolor[gray]{0.95}
 & Qwen3-235B-A22B & -0.010 & -0.003 \\
\rowcolor[gray]{0.95}
 & Gemini-3-Flash & 0.004 & -0.007 \\

\multirow[t]{4}{*}{GrailQA} & GPT-4.1-mini & -0.011 & -0.015 \\
 & GPT-5-mini & -0.035 & -0.019 \\
 & Qwen3-235B-A22B & \textbf{-0.073} & \textbf{-0.061} \\
 & Gemini-3-Flash & -0.039 & \textbf{-0.102} \\

\rowcolor[gray]{0.95}
\multirow[t]{4}{*}{WebQSP} & GPT-4.1-mini & 0.006 & -0.010 \\
\rowcolor[gray]{0.95}
 & GPT-5-mini & 0.011 & -0.013 \\
\rowcolor[gray]{0.95}
 & Qwen3-235B-A22B & 0.002 & -0.003 \\
\rowcolor[gray]{0.95}
 & Gemini-3-Flash & 0.000 & -0.002 \\

\multirow[t]{4}{*}{GraphQ} & GPT-4.1-mini & \textbf{-0.062} & \textbf{-0.056} \\
 & GPT-5-mini & -0.048 & \textbf{-0.062} \\
 & Qwen3-235B-A22B & \textbf{-0.098} & \textbf{-0.082} \\
 & Gemini-3-Flash & \textbf{-0.078} & \textbf{-0.178} \\

\rowcolor[gray]{0.95}
\multirow[t]{4}{*}{Mul. HotpotQA} & GPT-4.1-mini & \textbf{-0.107} & \textbf{-0.101} \\
\rowcolor[gray]{0.95}
 & GPT-5-mini & \textbf{-0.067} & 0.014 \\
\rowcolor[gray]{0.95}
 & Qwen3-235B-A22B & -0.045 & \textbf{-0.077} \\
\rowcolor[gray]{0.95}
 & Gemini-3-Flash & \textbf{-0.065} & \textbf{-0.190} \\
\bottomrule
\end{tabular}
\end{table}

To evaluate how SH and FH behave under different levels of tool robustness, we inject noise into tool execution. Specifically, for KBQA we disable automatic schema matching (forcing exact identifier use), and for multi-objective HotpotQA we restrict retrieval to top-1 (simulating low recall). Under both perturbations, tools are more likely to return empty results, which in turn necessitate replanning.

\paragraph{Result:} Table~\ref{tab:tool-robustness} shows that noise leads to comparable performance degradation for FH and SH. In many cases, the drop is less than 0.05 accuracy points, and the difference between the two paradigms is small, suggesting that both SH and FH handle execution failures reasonably well.

\paragraph{Does SH exhibit greater robustness in noisy environments?} No. Contrary to the expectation that immediate step-by-step feedback improves error recovery, SH shows no consistent advantage and is sometimes more sensitive to noise. This unexpected brittleness is most evident with Gemini-3-Flash. On GraphQ, turning off soft schema matching reduces SH accuracy by 0.178 points, in contrast to a 0.078-point drop for FH. On HotpotQA, restricting retrieval to top-1 reduces SH accuracy by 0.190 points, compared to a 0.065-point drop for FH. We analyze this behavior in more detail in the next section.

\begin{table}[t]
\centering
\caption{Fraction of task instances in which the agent made at least one repeated tool call (same tool call with identical arguments). SH exhibits substantially higher repetition rates than FH in several settings. Rates $\ge10\%$ are shown in bold.}
\label{tab:exp-repetition}
\begin{tabular}{llrr}
\toprule
Dataset & Model & \multicolumn{1}{c}{FH} & \multicolumn{1}{c}{SH} \\
\midrule
\rowcolor[gray]{0.95}
\multirow[t]{4}{*}{KQA Pro} & GPT-4.1-mini & 1.3\% & 2.1\% \\
\rowcolor[gray]{0.95}
 & GPT-5-mini & 0.7\% & 3.2\% \\
\rowcolor[gray]{0.95}
 & Qwen3-235B-A22B & 2.8\% & 1.6\% \\
\rowcolor[gray]{0.95}
 & Gemini-3-Flash & 0.1\% & \textbf{10.0\%} \\

\multirow[t]{4}{*}{GrailQA} & GPT-4.1-mini & 0.7\% & 1.1\% \\
 & GPT-5-mini & 0.3\% & 3.0\% \\
 & Qwen3-235B-A22B & 2.6\% & 2.2\% \\
 & Gemini-3-Flash & 1.9\% & \textbf{29.3\%} \\

\rowcolor[gray]{0.95}
\multirow[t]{4}{*}{WebQSP} & GPT-4.1-mini & 0.2\% & 0.4\% \\
\rowcolor[gray]{0.95}
 & GPT-5-mini & 0.2\% & 1.0\% \\
\rowcolor[gray]{0.95}
 & Qwen3-235B-A22B & 0.6\% & 0.6\% \\
\rowcolor[gray]{0.95}
 & Gemini-3-Flash & 0.6\% & 5.5\% \\

\multirow[t]{4}{*}{GraphQ} & GPT-4.1-mini & 0.8\% & 1.4\% \\
 & GPT-5-mini & 0.3\% & 3.1\% \\
 & Qwen3-235B-A22B & 1.6\% & 1.1\% \\
 & Gemini-3-Flash & 0.0\% & \textbf{35.8\%} \\

\rowcolor[gray]{0.95}
\multirow[t]{4}{*}{Mul. HotpotQA} & GPT-4.1-mini & 0.6\% & 0.6\% \\
\rowcolor[gray]{0.95}
 & GPT-5-mini & 6.3\% & \textbf{39.5\%} \\
\rowcolor[gray]{0.95}
 & Qwen3-235B-A22B & 1.5\% & 0.4\% \\
\rowcolor[gray]{0.95}
 & Gemini-3-Flash & 5.9\% & \textbf{45.5\%} \\
\bottomrule
\end{tabular}\end{table}

\begin{table}[t]
\centering
\small
\caption{Gemini-3-Flash under low-robustness retrieval (top-1) on multi-objective HotpotQA ($k=2$). This table focuses on one sub-question: ``Which director, John Schlesinger or Barbara Albert, was also a writer and film producer?'' Italic text shows the model’s internal reasoning. ``Error'' denotes execution failures. The feedback text returned to the agent is omitted for brevity. See Appendix \ref{app:case-study} for full trajectories.}
\label{tab:repetition-case-study}
\begin{minipage}{0.9\linewidth}
    \subcaption{SH trajectory: After failures, the agent eventually repeated an identical search call (shown in red).}
    \label{tab:repetition-case-study-sh}
    \begin{tabular}{p{1\textwidth}}
    \toprule
    \textit{I'm currently engaged in evaluating both John Schlesinger and Barbara Albert, determining if each functioned as a writer and producer. After gathering the facts, a comparison will readily reveal the answer. Parallel to this, ...} \\
    > search(Was John Schlesinger a writer and film producer?) \\
    \multicolumn{1}{r}{\color{red} $\to$ Error} \\
    \textit{I'm starting by investigating the first question. I've initiated Step 0: searching for John Schlesinger's roles. ...} \\
    > search(What were the occupations of director John Schlesinger?) \\
    \multicolumn{1}{r}{\color{red} $\to$ Error} \\
    \textit{I'm now formulating my search queries. I'll begin by querying about the occupations of John Schlesinger and Barbara Albert. ...} \\
    > search(What were the occupations of John Schlesinger?) \\
    \multicolumn{1}{r}{\color{red} $\to$ Error} \\
    \textit{I've begun looking into the roles of John Schlesinger and Barbara Albert. Initial searches for their occupations are underway. I'll need ...} \\
    > {\color{red} search(What were the occupations of John Schlesinger?)} [repeat] \\
    \bottomrule
    \end{tabular}
\end{minipage}\\ \vspace{1\baselineskip}
\begin{minipage}{0.9\linewidth}
    \subcaption{FH trajectory: Upon failures, the agent replans the remaining steps. It shifted focus to an alternative sub-goal, and intentionally refined the search query (bold).}
    \label{tab:repetition-case-study-fh}
    \begin{tabular}{p{1\textwidth}}
    \toprule
    \textit{I've initiated searches to determine if John Schlesinger and Barbara Albert were both writers and producers, as planned. Once I have those details, a comparison will be conducted to pinpoint the correct individual. I'm also ...} \\
    > [\$0 = search(Was John Schlesinger a writer and film producer?), \$1 = search(Was Barbara Albert a writer and film producer?), ...] \\
    \multicolumn{1}{r}{\color{red} $\to$ Error in \$1} \\
    \textit{I've restarted my search for John Schlesinger's occupations, focusing on ... Also, I'm planning to look into Barbara Albert's roles. ...} \\
    > [\$1 = search(What were the occupations of director John Schlesinger?), \$2 = search(What were the occupations of director Barbara Albert?), ...] \\
    \multicolumn{1}{r}{\color{red} $\to$ Error in \$1} \\
    \textit{\textbf{I'm now focusing on Barbara Albert's occupations,} and then refining how I search for John Schlesinger's roles. ...} \\
    > [\$2 = search(What are the occupations of Barbara Albert (director)?), \$3 = search(Was John Schlesinger a writer and film producer in addition to being a director?), ...] \\
    \multicolumn{1}{r}{\color{red} $\to$ Error in \$3} \\
    \textit{I'm focusing on John Schlesinger's occupations, trying to confirm if he was also a writer or producer. \textbf{I'll use a more specific search term this time around.} ...} \\
    > [\$4 = search(What were the professional roles of the English film director John Schlesinger?), ...] \\
    \bottomrule
    \end{tabular}
\end{minipage}

\end{table}

\subsection{Failure Mode Analysis: Repetitive Tool Calls}\label{sec:exp-failure-mode}

To better understand the unexpected brittleness of SH under some conditions, we identified a key failure mode, particularly in low-robustness tool settings: SH agents can get stuck in repetitive tool-call loops more often than FH agents. When a tool call fails or returns an empty result, models such as Gemini-3-Flash often issue the same tool call again in the next step with identical arguments, despite receiving failure feedback.

To quantify this behavior, we measured the fraction of task instances in which an agent made at least one repeated tool call, defined as invoking the same tool with identical arguments multiple times. Under the default high-robustness setting, Gemini-3-Flash in SH mode repeated tool calls in 30--45\% of instances on GrailQA and multi-objective HotpotQA (Table~\ref{tab:exp-repetition}), whereas FH reduced this rate to 1.9\% (GrailQA) and 5.9\% (multi-objective HotpotQA). The gap widened further in low-robustness settings: on multi-objective HotpotQA with top-1 retrieval, SH made repetitive calls in 66.9\% of instances, in contrast to 13.9\% for FH (see Table~\ref{tab:repetition-case-study} for a case study). We observed the same pattern, though less pronounced, with other LLM backends as well.

We conjecture that FH with lazy replanning is less prone to such local traps for two reasons. First, replanning is triggered only upon execution failure rather than at every step, which may reduce the chance of myopic retries. Second, when replanning is triggered, the model must generate a complete remaining plan toward the end goal, which may guide it to revise the overall strategy instead of repeating the same local action.

\section{Related Work}\label{sec:related-work}

We position our work within three strands of research: (1) comparative analyses of planning paradigms, (2) adaptive or hybrid planning strategies, and (3) agentic approaches for reasoning in data-centric tasks.

\subsection{Comparative Studies of Planning Paradigms}\label{sec:related-work-comparison}

Recent survey papers provide meta-analyses of the broad landscape of LLM-based task planning techniques~\cite{huang-etal-2024-understanding,li-2025-review,wei-etal-2025-plangenllms}. Closest to our analytic goals is the work of \citet{liu-etal-2025-select-decompose}, who depart from conceptual analysis to provide an empirical analysis of LLM task decomposition ($\approx$planning) across multiple domains. They show that the choice between Decomposition-First (analogous to FH) and Interleaved (analogous to SH) strategies is highly task-dependent, especially when computational efficiency is taken into account. This finding is particularly relevant given that step-by-step execution with eager monitoring remains a common default in low-level tool use under environmental uncertainty~\cite{kim-et-al-2024-llm,gonzalez-pumariega-etal-2025-robotouille,zhang-etal-2025-recap}.

Our work builds on \citeauthor{liu-etal-2025-select-decompose}'s insights but differs in three key aspects. First, we isolate \textit{planning horizon} as the key architectural feature and analyze its impact on both accuracy and cost at the level of tool execution. Second, while Liu et al. primarily evaluate pure LLM reasoning, we focus on \textit{data-centric tool calling}, which exhibits unique challenges (logic and vocabulary alignment). Third, we move beyond dataset-level comparisons by introducing an instance-level characterization of task difficulty to better understand when and why SH is (or is not) advantageous.

\subsection{Adaptive Planning Strategies}\label{sec:related-work-adaptive-planning}

As discussed in Section~\ref{sec:method-characterization}, existing planning techniques can be categorized into FH and SH, each with distinct characteristics. FH planning was common in early LLM planning work, motivated by multi-step reasoning paradigms such as Chain-of-Thought~\cite{wei-etal-2022-chain} and explicit plan-then-execute pipelines~\cite{wang-etal-2023-plan,lu-etal-2023-chameleon,schick-etal-2023-toolformer}. Although SH is now often viewed as the more robust default following strong results in prior work~\cite{inaba-etal-2023-multitool,yao-etal-2023-react}, FH continues to motivate active research, including techniques that verify or optimize a full plan before execution~\cite{li-etal-2025-agent,lee-etal-2025-veriplan}, as well as approaches that prioritize lower computational cost~\cite{xu-etal-2023-rewoo}.

To combine their strengths, recent work has explored strategies that integrate upfront global planning with iterative local planning. One prominent direction is adaptive workflows, where agents dynamically consolidate low-level actions into a single step~\cite{wang-etal-2024-executable} or begin with a global plan and switch to step-wise planning when execution fails or task complexity demands it~\cite{prasad-etal-2024-adapt,choi-et-al-2025-reactree}. Another direction is component optimization, where the roles of global planner and local executor are optimized separately using specialized fine-tuning~\cite{erdogan-etal-2025-plan} or exemplar retrieval~\cite{kim-etal-2024-rada}. Practical systems (e.g., LLM-powered coding agents) also increasingly support mixed planning behaviors such as upfront task decomposition and batched tool execution~\cite{anthropic-2025-ptc,cloudflare-2025-codemode}.

Yet, the design of these integrated systems remains largely heuristic, with limited systematic understanding of exactly which cues should trigger strategy switching. Our work does not introduce a new systems paradigm. Rather, we provide a controlled empirical comparison of planning horizon in data-centric tool calling to clarify when FH-style behavior can be a cost-effective option. This insight motivates future work on principled routing and strategy selection grounded in measurable task properties rather than trial-and-error adaptation.

\subsection{Agentic Approaches to Data-centric Tasks}\label{sec:related-work-data-agents}

Traditionally, data-centric tasks like KBQA and Text-to-SQL were solved by specialized semantic parsers. Recently, significant attention has shifted toward agentic data analytics~\cite{testini-etal-2025-measuring,chen-etal-2025-large}. In these frameworks, LLMs interact with environments to access and manipulate data for complex tasks. Frameworks like Middleware~\cite{gu-etal-2024-middleware} and Interactive-T2S~\cite{xiong-etal-2025-multi} enable LLM agents to gain schema awareness through interaction, addressing the hallucination problems inherent in static query generation.

Tight step-wise feedback can be essential when tasks require exploratory schema discovery or when the environment is highly uncertain. However, for data-centric tasks with relatively stable tool and data semantics, our analysis suggests that SH is not always the most cost-effective default. In production environments, latency and token budgets are often primary constraints. Our findings therefore offer a practical perspective on designing efficient data-centric agents.

\section{Conclusion}\label{sec:conclusion}

We evaluate whether LLM agents solving data-centric tool-calling tasks require step-wise planning with eager monitoring in the settings we study. We isolate \emph{planning horizon} as the key architectural choice, comparing single-step horizon (SH) planning with full-horizon (FH) planning under \emph{lazy} monitoring. To make this comparison controlled and fine-grained, we characterize task difficulty along \emph{topological complexity} (execution-graph depth and breadth) and \emph{tool robustness} (tolerance to imperfect inputs), and we evaluate across KBQA and multi-hop QA with multiple LLMs.

Despite the common assumption favoring SH, we find no consistent accuracy advantage for SH over FH, nor evidence that SH is more robust to increased topological complexity or noisy tool behavior. Overall, these results suggest that, in the \emph{well-defined} data-centric tool-use settings studied here, FH with lazy replanning can match SH while using substantially fewer tokens, while SH may remain preferable in more exploratory or highly uncertain environments.

These findings open several directions for future work. An important next step is to evaluate and extend these methods in more complicated and real-world scenarios, including domain-specific or proprietary data. Our results also motivate developing hybrid methods that build on the complementary strengths of FH and SH.

\begin{acks}
We thank the anonymous reviewers and our colleagues at Megagon Labs for their valuable feedback and discussions.
\end{acks}

\bibliographystyle{ACM-Reference-Format}
\bibliography{custom}

\appendix
\onecolumn

\section{Artifact Appendix}

Our GitHub repository (\url{https://github.com/megagonlabs/cais26-planning-horizon}), includes
source code, configuration files, scripts, and analysis notebooks. Datasets
and raw experiment outputs are not bundled. In this section, we provide a brief introduction to our codebase.

\subsection{Key Results and Reproduction Summary}

The main claim of the paper is
that, \textbf{for well-defined data-centric tasks, a full-horizon (FH) planner with
lazy monitoring can match a single-step horizon (SH) planner while avoiding
much of the overhead from constant feedback integration.} The table below summarizes
how the main findings are supported by our software.

\begin{enumerate}
    \item \textbf{FH matches SH accuracy across datasets and backbone models while FH uses fewer tokens than SH.}
    (Table~\ref{tab:main-results} in Section~\ref{sec:exp-main})
    \begin{itemize}
        \item Run \texttt{scripts/batch/batch\_exp\_*\_hydra.sh}
        wrappers, then \texttt{scripts/batch/batch\_postprocess\_main\_results.sh} for comparing accuracy and token usage between SH and FH across datasets and models.
        \item Details:
        \texttt{docs/walkthrough.md}, \texttt{README.md}
    \end{itemize}

    \item \textbf{Topological complexity hurts SH and FH \textit{in similar ways.}}
    (Table~\ref{tab:glm-coefficients} in Section~\ref{sec:exp-topological-complexity})
    \begin{itemize}
        \item Run
        \texttt{notebooks/sec4p3\_topological-complexity-analysis.ipynb} to analyze the results from (1).
        \item Details: \texttt{docs/walkthrough.md}
    \end{itemize}

    \item \textbf{SH shows \textit{no consistent robustness advantage} in noisy tool settings.}
    (Table~\ref{tab:tool-robustness} in Section~\ref{sec:exp-tool-robustness})
    \begin{itemize}
        \item Run
        \texttt{notebooks/sec4p4\_tool-robustness.ipynb} to analyze the results from (1).
        \item Details: \texttt{docs/walkthrough.md}
    \end{itemize}

    \item \textbf{SH repeats tool calls more often after failures.}
    (Section~\ref{sec:exp-failure-mode})
    \begin{itemize}
        \item Run
        \texttt{notebooks/sec4p5\_repetitive-tool-calls.ipynb} to analyze the results from (1).
        \item Details: \texttt{docs/walkthrough.md}
    \end{itemize}
\end{enumerate}

See Appendix~\ref{sec:entry-points} for more details.

\subsection{Requirements}

Requirements vary by experiment type. See \texttt{README.md} for
details.

\begin{itemize}
    \item \textbf{Software:} Python 3.11+ and \texttt{uv} are required. Most
    dependencies can be installed with \texttt{uv}. API keys for the
    selected LLM providers (e.g., OpenAI) must be set in the repository-root \texttt{.env}
    file. Atomic KBQA also needs Virtuoso.
    \item \textbf{Hardware:} The experiment pipeline needs a GPU. Hardware
    needs depend on the experiment track. RAM and disk must fit the selected
    knowledge base or Wikipedia-based corpus. Multi-objective HotpotQA uses
    large Pyserini indexes and may need more RAM during retrieval.
    Freebase-backed Atomic KBQA runs recommend 100 GB+ RAM.
    \item \textbf{Network:} Internet access is needed for LLM APIs and for
    first-time index downloads.
\end{itemize}

\subsection{Setup Guide}

\noindent\textbf{Read \texttt{docs/setup/code.md} and
\texttt{docs/setup/data.md} first.} Then use the dataset READMEs under
\texttt{data/} for track-specific setup. If you want the lightest path first,
start with KQA Pro, then move to the larger tracks.

\subsection{Execution Entry Points}\label{sec:entry-points}

This section briefly describes the entry points to reproduce the key experimental results. \textbf{\texttt{docs/walkthrough.md} is the main guide.}

\subsubsection{Main Experiment Workflow (\S\ref{sec:exp-main})}

For Tables~\ref{tab:main-results} and \ref{tab:tool-robustness}, run \texttt{scripts/batch/batch\_exp\_*\_hydra.sh} wrappers for the main experiments. The results can be evaluated and aggregated by \texttt{batch\_postprocess\_main\_results.sh}. This workflow supports the findings on accuracy parity and token reduction. The output files are used in the subsequent post-hoc analyses.

\subsubsection{Topological Complexity Analysis (\S\ref{sec:exp-topological-complexity})}

\texttt{notebooks/sec4p3\_topological-complexity-analysis.ipynb} supports the
finding that greater execution-graph depth and breadth hurt both planners in
similar ways, without a clear SH-specific advantage. It produces the
logistic-regression coefficients used for Table~\ref{tab:glm-coefficients}.

\subsubsection{Tool Robustness Analysis (\S\ref{sec:exp-tool-robustness})}

\texttt{notebooks/sec4p4\_tool-robustness.ipynb} supports the finding that
both SH and FH planners degrade by similar amounts when tool robustness is reduced, with no consistent SH advantage. It produces the accuracy-delta tables used in Section~\ref{sec:tool-robustness}.

\subsubsection{Repetitive Tool Call Analysis (\S\ref{sec:exp-failure-mode})}

\texttt{notebooks/sec4p5\_repetitive-tool-calls.ipynb} supports the finding
that SH repeats tool calls much more often than FH after failures. It produces
the repetition-rate values mentioned in Section~\ref{sec:exp-failure-mode}.

\subsubsection{Lightest Runnable Example for Sanity-Check}

\texttt{scripts/example\_run\_kqa\_pro.py} is the smallest end-to-end run in the repository. It takes a user query as a commandline argument and shows how SH/FH agents answer it using KoPL.

\subsection{Cautions}

A few practical issues are worth knowing before you run the code.

\begin{itemize}
    \item \textbf{Nondeterminism:} LLM API runs are not fully deterministic.
    Reproduced numbers should be close to the paper, but exact
    score-by-score matches are not expected.
    \item \textbf{Heavy setup:} Atomic KBQA needs a local Virtuoso and
    Freebase setup. Multi-objective HotpotQA downloads large Pyserini indexes
    on first use. These datasets need more setup time and more memory.
    \item \textbf{Execution time:} For the representative OpenAI settings
    reported in \texttt{README.md}, one trial is on the order of about 2 to
    64 hours of total runtime, depending on track, planner, and model.
    Wall-clock time can be lower with parallel execution.
\end{itemize}

\subsection{Recommendations}

The paths below are the easiest ways to approach the artifact.

\begin{itemize}
    \item \textbf{Start small:} Start with KQA Pro if you want the lightest runnable path. Use the smaller-run options in the batch scripts for sanity-check.
    \item \textbf{Use parallel execution when possible:} If multiple GPUs are
    available, use the parallel settings. (See the walkthrough document.)
\end{itemize}

\section{Implementation Details}\label{app:implementation-details}

This section describes the details of tasks~(\ref{app:task-details}), planners~(\ref{app:planner-details}), and evaluation~(\ref{app:evaluation-details}).

\subsection{Task Details}\label{app:task-details}

Our study uses KQA Pro, Atomic KBQA datasets (GrailQA, WebQSP, and GraphQ), and the multi-objective version of HotpotQA. They undergo distinct preprocessing and are handled by different tool sets. We describe these details below.

\subsubsection{KQA Pro}

KQA Pro~\cite{cao-etal-2022-kqa} is a dataset for QA over a sampled subset of Wikidata~\cite{vrandecic-and-krotzsch-2014-wikidata}. It features approximately 120k examples paired with ground-truth reasoning programs in KoPL (Knowledge-oriented Programming Language).

\begin{table*}[t]
    \centering
    \small
    \begin{tabular}{lp{0.4\linewidth}p{0.4\linewidth}}
    \toprule
    Category & KoPL Tools & Description \\
    \midrule
    Lookup & \texttt{FindAll,} \texttt{Find} & Look up and return an entity set \\
    Filtering & \texttt{FilterConcept}, \texttt{FilterStr}, \texttt{FilterNum}, \texttt{FilterYear}, \texttt{FilterDate}, \texttt{QFilterStr}, \texttt{QFilterNum}, \texttt{QFilterYear}, \texttt{QFilterDate} & Filter the input entity set by a specified condition \\
    Traversal & \texttt{Relate} & Traverse to connected an entity set via a specified relation \\
    Logic & \texttt{And}, \texttt{Or} & Take the intersection or union of two input entity sets \\
    Aggregation & \texttt{Count}, \texttt{SelectAmong}, \texttt{SelectBetween} & Count an entity set or select an entity by a specified attribute \\
    Verification & \texttt{VerifyStr}, \texttt{VerifyNum}, \texttt{VerifyYear}, \texttt{VeryfyDate} & Verify if the specified attribute of the input entity set meets a criteria \\
    Projection & \texttt{QueryName}, \texttt{QueryRelation}, \texttt{QueryAttr}, \texttt{QueryAttrUnderCondition}, \texttt{QueryAttrQualifier}, \texttt{QueryRelationQualifier} & Access a specified value of the input entity set \\
    \bottomrule
    \end{tabular}
    \caption{KoPL tools defined by \citet{cao-etal-2022-kqa}.}
    \label{tab:kopl-tools}
\end{table*}

\begin{lstlisting}[style=booktabsstyle, caption={Schema grounding for KQA Pro}, label={prmt:schema-grounding-kqa-pro}]
[system] (FH)
Use the available tools to answer the user's question step by step, generating the entire plan. Use $i to refer to the output of step i (0-indexed).

**Tools**
{tool_definitions}

You can access the knowledge base through the provided tools.

**Examples**
{demonstrations}

[system] (SH)
Use the available tools to answer the user's question step by step, generating one action at a time. Use $i to refer to the output of step i (0-indexed).

(...same as FH)

[user]
Question: {question}
\end{lstlisting}

\paragraph{Tools:} The agent has access to 27 tools based on KoPL functions~(Table~\ref{tab:kopl-tools}).\footnote{\url{https://github.com/THU-KEG/KoPL}} We derived the ground-truth execution graph of KoPL tools based on the program annotations in the original data. When parameters (concept, attribute, or relation) specified by the planner are not found in the KB schema, we perform soft matching based on embedding similarity and an LLM. We first retrieve 10 similar values based on \texttt{bge-base-en-v1.5} embedding similarity, followed by evaluation with \texttt{gpt-4.1-mini-2025-04-18} (see Prompt~\ref{prmt:schema-grounding-kqa-pro} for prompts). If no valid candidate is found, the tool returns the top-10 similar candidates as feedback. For the controlled experiments regarding tool robustness~(\S\ref{sec:exp-tool-robustness}), we disable this soft matching and only return the top-1 candidate to simulate tools with low robustness.

\paragraph{Preprocessing:} We sampled approximately 1,000 examples each from the training and development splits of KQA Pro, resulting in new training and test sets with balanced step lengths.
\begin{enumerate}
    \item \textbf{Validation:} We filtered examples to ensure the provided KoPL program executes without error and matches the ground-truth answer.
    \item \textbf{Plan DAG Annotation:} KoPL programs are represented as trees. For our analysis, we converted them into DAGs by merging identical steps.
    \item \textbf{Sampling:} We balanced the test set across 5 complexity bins based on the number of DAG nodes (1-3, 4-5, 6-7, 8-9, 10+).
    \item \textbf{In-context Demonstration:} For each task instance, we retrieved 10 similar examples from held-out data as in-context demonstrations using \texttt{bge-base-en-v1.5} embeddings of the question text.
\end{enumerate}

\subsubsection{Atomic KBQA (GrailQA, WebQSP, GraphQ)}

GrailQA~\cite{gu-etal-2021-beyond}, WebQSP~\cite{yih-etal-2016-value}, and GraphQ~\cite{su-etal-2016-generating} contain QA instances over the full Freebase knowledge base.\footnote{We used the resources available at \url{https://github.com/dki-lab/Freebase-Setup} for running the Freebase server locally.} Each dataset contains annotations of logical forms called S-expressions~\cite{gu-etal-2021-beyond}, which can be deterministically converted into executable SPARQL queries. We refer to these datasets collectively as \textit{Atomic KBQA} datasets because we employ Atomic Query Tools (below) for them.

\begin{table*}
    \centering
    \small
    \begin{tabular}{llp{0.4\textwidth}}
    \toprule
    Atomic Query Tools & S-expression & Description \\
    \midrule
    \texttt{Extract\_entity(input)} & \texttt{START} & Resolve an input (entity mention, entity class, or literal) to Freebase entity IDs or a typed literal. \\
    \texttt{Find\_relation(relation, direction, target)} & \texttt{JOIN} & Find entities that point to the given target entities via the specified Freebase relation (?x -relation-> target) \\
    \texttt{Merge(input1, input2)} & \texttt{AND} & Compute set intersection of two entity sets \\
    \texttt{Order(mode, input, property)} & \texttt{ARGMIN}/\texttt{ARGMAX} & Find entities with maximum or minimum property value (mode = \{argmin, argmax\}) \\
    \texttt{Compare(operator, property, literal)} & \texttt{LT}/\texttt{LE}/\texttt{GT}/\texttt{GE} & Find entities whose property compares to a literal (operator = \{<, <=, >, >=\}) \\
    \texttt{Time\_constraint(input, relation, literal)} & \texttt{TC} & Filter an input entity set by equality of a temporal property to a year, or 'NOW'. \\
    \texttt{Count(input)} & \texttt{COUNT} & Count the number of input entities \\
    \bottomrule
    \end{tabular}
    \caption{Atomic query tools defined by~\citet{luo-etal-2025-kbqa} with modifications, and their correspondence to functions in S-expression~\cite{gu-etal-2021-beyond}.}
    \label{tab:atomic-query-tools}
\end{table*}

\paragraph{Tools:} We implemented seven \textit{Atomic Query Tools}~\cite{luo-etal-2025-kbqa}\footnote{\url{https://github.com/LHRLAB/KBQA-o1}} with slight modifications~(Table~\ref{tab:atomic-query-tools}). These tools wrap functions in S-expressions. Chains of function calls can be converted into S-expressions, and subsequently to SPARQL queries. We derived the ground-truth execution graph based on the S-expression annotations. Similar to KoPL tools, we perform soft schema grounding using embeddings and an LLM when the parameters specified by the planner are not found.

\paragraph{Preprocessing:} We sampled roughly 1,000 examples from the three datasets for creating training and test sets with balanced step lengths.
\begin{enumerate}
    \item \textbf{Deduplication:} These datasets often contain multiple variations derived from the same logical form template. We grouped examples by unique S-expression to avoid bias toward common templates.
    \item \textbf{Validation:} We ran SPARQL queries generated from the ground-truth S-expressions and filtered out those that returned answers differing from the ground truth.
    \item \textbf{Plan DAG Annotation:} We parsed linear function lists into DAGs based on variable usage.
    \item \textbf{Sampling:} We balanced the dataset across 4 complexity bins (1-3, 4-5, 6-7, 8+). \\
    \item \textbf{Demonstration Retrieval:} We retrieved 10 similar examples for each query using question embeddings, similar to the KQA Pro procedure.
\end{enumerate}

\subsubsection{Multi-Objective HotpotQA}

HotpotQA~\cite{yang-etal-2018-hotpotqa} is widely used in existing work on LLM agents. To analyze the effect of structural complexity, we followed \citet{zhou-etal-2025-mem1} and synthesized multi-objective tasks by combining $k$ independent HotpotQA questions ($k = 2, \dots, 5$) in addition to the original HotpotQA instances ($k=1$).

\paragraph{Tools:} We implemented LLM sub-agents~\cite{kim-etal-2024-husky} as tools. We used \texttt{gpt-4.1-mini-2025-04-18} as a backbone.
\begin{enumerate}
    \item \texttt{search(question)}: Answers a single-hop question based on evidence retrieved from English Wikipedia\footnote{We used the 2017-10-01 dump of the first paragraphs released by \citet{yang-etal-2018-hotpotqa}.} using the pre-built Pyserini index with \texttt{bge-base-en-v1.5}~\cite{lin-etal-2021-pyserini}. By default, we retrieve the top-10 relevant paragraphs for the search query. In the controlled experiments~(\S\ref{sec:exp-tool-robustness}), we limit this to the top-1 result to simulate a less robust search tool.
    \item \texttt{commonsense(question)}: Performs logic/comparison using the internal knowledge of the LLM.
\end{enumerate}

\begin{lstlisting}[style=booktabsstyle, caption={Validation of HotpotQA questions}, label={prmt:hotpotqa-validation}]
[system]
You are validating whether a HotpotQA bridge question follows a valid linear reasoning structure. Judge only from the wording of the question; do not use outside knowledge or supporting facts. Decompose the question into steps, state if any step yields multiple candidates (a set), and then output a single JSON object.

Core definitions
- Linear chain (valid): A strictly sequential chain of 2+ steps where each step yields exactly one intermediate entity/property that directly feeds the next step. Direct property lookup or a single yes/no verification on that single entity is fine.
- Branching/set filtering (invalid): Any step produces multiple candidates that must be checked/filtered individually (set operations), or the question asks for intersections/commonalities/comparisons across multiple entities.

Conservative uniqueness policy (do not assume):
- Do not infer uniqueness from plausibility or typical world facts. If the wording does not make uniqueness explicit, treat the step as set-producing -> invalid.
- Roles that vary over time (e.g., "the president of [org]", "the coach of [team]") are non-unique unless the question specifies a time/tenure (e.g., a year/season/ordinal/current).
- Combining multiple descriptors (e.g., "Austrian forest caretaker, naturalist, pseudoscientist") does not guarantee uniqueness; still treat as potentially multiple unless uniquely pinned.

Valid patterns (usually unique by wording)
- Specific titled work -> role/property (e.g., "the director of [titled work]", "the vocalist on '[song]'").
- Definite, singular roles tied to a specific proper noun with an explicit time/ordinal (e.g., "the head coach of [team] in [year]", "the 42nd president of [country]").
- Definitional superlatives that denote a single item by definition (e.g., "the capital of [country]", "the largest city in [county]").

Invalid patterns (must label invalid)
- Set-producing first step (guests, stars, cast members, contestants, authors, films featuring X, etc.) followed by filtering.
- Parallel/compound questions seeking two independent facts or commonalities (e.g., "What do X and Y have in common?", "Which star of A was also in B?").
- Ambiguous cardinality or vague qualifiers (e.g., "former", "long-time") without time bounds; treat as set-producing.
- Any step where uniqueness across time is not fixed by the wording.

Output requirements
- Return only a single JSON object with keys:
    - "reasoning": 2-4 concise sentences that enumerate the steps (e.g., "Step 1... Step 2...") and explicitly state where (if anywhere) branching occurs.
    - "is_valid": true if linear (single-path), false if branching/set/parallel or if only a single-hop lookup.
- Do not answer the original question. Do not include extra keys, disclaimers, or formatting.

Always think step by step based solely on the question text, err on the side of treating ambiguous steps as branching, indicate whether branching occurs and where, and then provide the final JSON.

[user]
Question: {question}

Is the reasoning structure of this question valid according to the criteria provided? Decompose the question into stepwise reasoning (label Step 1, Step 2, etc.), indicate if and where branching (set operations or multiple candidate entities) is required, and return only the required JSON object.
\end{lstlisting}

\begin{lstlisting}[style=booktabsstyle, caption={DAG annotation of HotpotQA questions (Comparison:1/2)}, label={prmt:hotpotqa-dag-annotation-1}]
[system]
You're tasked with decomposing multi-hop "comparison" questions into exactly two parallel `search` steps followed by a single `reasoning` step-never fewer.

For every comparison question that requires choosing between Entity A and Entity B based on an attribute, follow this strict pattern:
- **Node 0:** Retrieve Entity A's attribute with `search`.
- **Node 1:** Retrieve Entity B's attribute with `search`.
- **Node 2:** Use `reasoning` to compare $0 vs $1 and output the required entity/value.

A comparison question **must** include one reasoning step that explicitly compares the two retrieved values. Annotate every example accordingly.

## Context
- **Goal:** Annotate multi-hop QA examples with ground-truth reasoning plans as a Directed Acyclic Graph (DAG). Your outputs must illustrate the required multi-step pattern.
- **Tools Available:**
    - `search(query)`: Retrieves facts (returns a string).
    - `reasoning(instruction)`: Applies simple logic (comparison, filtering, conditionals) on strings.
- **Annotator Role:** The answer and supporting facts are provided for your annotation. The QA system you annotate for cannot see them; avoid any leakage or hinting in your node inputs.

## Required "Comparison" Pattern
- For all "comparison" questions:
    - The DAG must contain exactly:
        - Two `search` nodes with no dependencies (parallel retrieval).
        - One `reasoning` node that depends on both searches.
    - The two `search` queries must be parallel in phrasing (same attribute asked for both entities) so outputs are directly comparable.
    - The final decision must be made in the `reasoning` node, not via a third `search`.

## Process
1. **Rephrase:** Restate the input question in clear, natural English, keeping all constraints.
2. **Construct DAG:** Break down the question into a minimal, strictly ordered DAG using the above rules and the tools provided.

## Node Construction Rules
- **Self-Contained:** Each node's `input` is a clear, standalone sentence (replace `$i` with the literal value).
- **Natural Embedding:** `$i` appears as though the entity or value is being asked about, not referenced. Never meta-phrase.
- **Literal Use:** Treat `$i` as a value to inquire about-not a doc/source.
- **Conciseness:** Each node contains exactly **one** input sentence.
- **No Redundant Search:** Never search for information already stated in the question; instead, use it in a reasoning node if involved.

## Reasoning Node Requirements (Comparison)
- Must be exactly one sentence.
- Must explicitly name both entities and reference both retrieved values ($0 and $1).
- Must state what to return (e.g., "output PersonA").
- Use a direct comparative construction; avoid awkward roles for $i (e.g., do not write "If $0 shows...").
- Template: "[EntityA] has [attribute] $0, and [EntityB] has [attribute] $1; based on the question's criterion, output the correct entity."

## Output Format
Return a single flat JSON object, no Markdown, no comments:
- `rephrased_question` (string)
- `dag`: List of nodes:
    - `function`: "search" or "reasoning"
    - `dependencies`: List of integers (node indices)
    - `input`: String (standalone, with `$i` references as needed)
\end{lstlisting}

\begin{lstlisting}[style=booktabsstyle, caption={DAG annotation of HotpotQA questions (Comparison:2/2)}, label={prmt:hotpotqa-validation-2}]
[system] (cont'd)
# Example

{examples}

# Notes
- Do not collapse comparison questions into one search.
- Do not add extra search steps after the two parallel searches.
- Always retrieve the same attribute for both entities with parallel phrasing.
- Always end with exactly one reasoning node that makes the selection and states what to output.
- Never reference the answer or supporting facts in any node input fields.

[user]
Generate the reasoning DAG for this question.

Question: {question}
Answer: {answer}
Supporting Facts: {supporting_facts}

**Notes**
- Do not collapse comparison questions into one search.
- Do not add extra search steps after the two parallel searches.
- Always retrieve the same attribute for both entities with parallel phrasing.
- Always end with exactly one reasoning node that makes the selection and states what to output.
- Never reference the answer or supporting facts in any node input fields.
\end{lstlisting}

\paragraph{Preprocessing:} We sampled examples from the training and development sets of HotpotQA and synthesized new training and test splits with multi-objective questions ($k=1, \cdots, 5$).
\begin{enumerate}
    \item \textbf{Validation:} As the original HotpotQA questions were generated based on pre-selected pairs of Wikipedia paragraphs, some questions, particularly bridge questions, are not suitable for tool calling settings.\footnote{For example, some bridge questions involve set questions with very large number of intermediate answers (e.g., all players in a specific team)}. To control reasoning structures, we used \texttt{gpt-5-mini-2025-08-07} to select bridge question that involve single entities as pivot answers. See Prompt~\ref{prmt:hotpotqa-validation}.
    \item \textbf{Sampling:} We sampled 200 examples uniformly across combinations of bridge/comparison types for each $k$, resulting in 1,000 QA pairs for each of the training and test splits.
    \item \textbf{Plan DAG Annotation:} We used \texttt{gpt-5.2-2025-12-11} with medium reasoning effort to generate ground-truth DAGs based on the gold supporting paragraphs for each sub-question. The annotation was straightforward for most questions (bridge questions generate \texttt{search} $\to$ \texttt{search}, and comparison questions generate two independent \texttt{search} steps followed by a comparison step with \texttt{commonsense}), but some bridge questions involved additional reasoning steps. We also generated \texttt{question} texts of tool calls to be used in in-context demonstrations. Prompt~\ref{prmt:hotpotqa-dag-annotation-1} shows the instructions to convert comparison questions. We used similar instructions for bridge questions.
    \item \textbf{Demonstration Retrieval:} We retrieved 10 demonstration examples with the same $k$ for each query using question embeddings.
\end{enumerate}

\subsection{Planner Details}\label{app:planner-details}

\begin{lstlisting}[style=booktabsstyle, caption={Planning prompts for KQA Pro. The glossary was defined by \citet{cao-etal-2022-kqa}.}, label={prmt:kqa-pro}]
[system] (FH)
Use the available tools to answer the user's question step by step, generating the entire plan. Use $i to refer to the output of step i (0-indexed).

**Tools**
{tool_definitions}

You can access the knowledge base through the provided tools.

**Glossary of KB concepts:**
- Entity: The most basic item in KB.
- Concept: The abstraction of a set of entities, e.g., basketball player.
- Relation: The link between entities or concepts. Entities are linked to concepts via the relation instance of. Concepts are organized into a tree structure via relation subclass of.
- Attribute: The literal information of an entity. An attribute has a key and a value, which is one of four types: string, number, date, and year. The number value can have an extra unit, e.g., 206 centimetre.
- Relational knowledge: The triple with form (entity, relation, entity), e.g., (LeBron James Jr., father, LeBron James).
- Literal knowledge: The triple with form (entity, attribute key, attribute value), e.g., (LeBron James, height, 206 centimetre).
- Qualifier knowledge: The triple whose head is a relational or literal triple, e.g., ((LeBron James, drafted by, Cleveland Cavaliers), point in time, 2003). A qualifier also has a key and a value.

**Examples**
{demonstrations}

[system] (SH)
Use the available tools to answer the user's question step by step, generating one action at a time. Use $i to refer to the output of step i (0-indexed).

(...same as FH)

[user]
Question: {question}
\end{lstlisting}

\begin{lstlisting}[style=booktabsstyle, caption={Planning prompts for multi-objective HotpotQA}, label={prmt:multiobj-hotpotqa}]
[system] (FH)
Use the available tools to answer the user's question step by step, generating the entire plan. Use $i to refer to the output of step i (0-indexed).

**Tools**
{tool_definitions}

Use the available tools to answer the user's question step by step

**Examples**
{demonstrations}

[system] (SH)
Use the available tools to answer the user's question step by step, generating one action at a time. Use $i to refer to the output of step i (0-indexed).

(...same as FH)

[user]
Question: {question}
\end{lstlisting}

\begin{lstlisting}[style=booktabsstyle, caption={Replanning prompt (FH). This user message is appended to the message list. \texttt{\{execution\_result\}} is replaced with a list of executed actions and their results.}, label={prmt:FH-replanning}]

[user]
Execution Result:
{execution_result}

The last step returned an error or no results. Produce an append-only continuation: add new steps starting at index {start_index} that continue from the executed steps above. Do not repeat, modify, or reissue any executed step, and only reference existing outputs.

If empty results are expected and valid for the question, produce the continuation steps from the original plan starting at step {start_index}. Otherwise, add corrective steps starting at {start_index} (e.g., adjust tool parameters, or try alternative tools) and proceed toward the answer.
\end{lstlisting}

For both FH and SH planners, we use minimal prompts containing task instructions, tool definitions (JSON string)\footnote{We tested other formats like TypeScript and Markdown but found no significant difference. The JSON format generally performed well.}, and demonstration examples~(Prompts~\ref{prmt:atomic-kbqa} and \ref{prmt:multiobj-hotpotqa}). We include 10 in-context demonstrations retrieved via \texttt{BAAI/bge-base-en-v1.5} embeddings. FH generates the full list of steps. If execution fails, the agent receives the error and replans up to 8 times~(Prompt~\ref{prmt:FH-replanning}). SH generates one step, receives the observation, and generates the next step. Both FH and SH agents are allowed to issue up to 30 tool calls.

We use structured output functionality to constrain LLM generation to valid JSON structures representing a sequence of tool calls. Specifically, we use the \texttt{anyOf} keyword in the JSON Schema\footnote{\url{https://json-schema.org/}} to limit the vocabulary of action names and their parameter names. When the planner generates invalid parameter values (e.g., references to non-existent steps), we prompt it to retry with a simple error message: ``Your response is in an invalid format. Please read the instructions carefully and try again,'' allowing up to 8 retries.

We use the following backbone models via public APIs (OpenAI API for GPT, Fireworks API for Qwen, and Vertex AI API for Gemini).
\begin{itemize}
    \item \texttt{gpt-4.1-mini-2024-07-18} (Temperature 0)
    \item \texttt{gpt-5-mini-2025-08-07} (Temperature 1, Medium reasoning effort)
    \item \texttt{qwen3-235b-a22b-thinking-2507} (Temperature 0)
    \item \texttt{gemini-3-flash-preview} (Temperature 0, Medium reasoning effort)
\end{itemize}
We set the context limit to be sufficiently large (10k) and use default values for other parameters.

\subsection{Evaluation Details}\label{app:evaluation-details}

\begin{lstlisting}[style=booktabsstyle, caption={Evaluation prompt for KBQA tasks}, label={prmt:evaluation-kbqa}]
[system] (KBQA)

You are a strict, impartial grader of answer correctness.

**Goal:** Compare System Output to Correct Answer only. Do not use outside knowledge; treat Correct Answer as ground truth.

**Output:** Reply with exactly one label (no quotes, no punctuation, no extra text):
- correct
- partially_correct
- incorrect
- refusal/unsure

**Scoring rules:**
1) Non-answer -> refusal/unsure:
    - System Output refuses, asks for clarification, says "unknown"/"no information", or otherwise does not contain an answer value.

2) Exact match -> correct:
    - System Output conveys the same answer value(s) as in Correct Answer, and no additional answer values.
    - Treat lists as sets: ignore order, casing, surrounding filler, and duplicate repetitions of the same correct value(s).
    - Ignore purely explanatory filler (e.g., "the answer is ...").

3) Partial overlap -> partially_correct:
    - At least one expected answer value appears in the System Output, but it is missing any other required values and/or includes extra answer values not in the Correct Answer.

4) Mismatch -> incorrect:
    - None of the expected answer values appear in the System Output, or any provided value contradicts the Correct Answer.

**Definitions and matching guidance:**
- Answer value: an entity (ID or name), number, date, or other atomic item. Treat comma/semicolon/newline/bulleted lists and clear conjunctions as multiple values.
- Entity matching: "m.xxxxx (Name)"/"Qxxxxx (Name)" matches if either the ID or the Name appears in System Output (case-insensitive).
- Numeric/date matching: require exact equality unless Correct Answer explicitly lists multiple acceptable values. Rounded/truncated numbers are incorrect when Correct Answer provides a more precise value.
- Type/Hierarchy: Do not give credit for broader/narrower/related categories, roles, or specializations (e.g., hypernym/hyponym). Only the exact expected value counts.
- Strict comparison: Do not use the question text or outside knowledge to infer equivalence. Compare the System Output strictly to the Correct Answer text.
- Extra information:
    - Extra explanatory text may be ignored.
    - Extra answer values (additional entities/values) make the result partially_correct.

[user]
Question: {question}
Correct Answer: {correct_answer}
System Output: {system_output}
\end{lstlisting}

\begin{lstlisting}[style=booktabsstyle, caption={Evaluation prompt for multi-objective HotpotQA}, label={prmt:evaluation-multiobj-hotpotqa}]
[system]
You are a strict, impartial grader of answer correctness.

**Goal:** Compare System Output to Correct Answer only. Do not use outside knowledge; treat Correct Answer as ground truth.

**Output:** Reply with exactly one label (no quotes, no punctuation, no extra text):
- correct
- partially_correct
- incorrect
- refusal/unsure

**Scoring rules:**
1) Non-answer -> refusal/unsure:
    - System Output refuses, asks for clarification, says "unknown"/"no information", or otherwise does not contain an answer value.

2) Exact match -> correct:
    - The System Output conveys exactly the same answer value(s) as the Correct Answer and no additional distinct answer values, in the same order for multi-part answers.
    - Ignore casing, articles, punctuation/spacing, and duplicate repetitions.
    - Ignore explanatory or descriptive context that does not add distinct answer values (e.g., prose around the answer, unit qualifiers, city+state after a city, manufacturer prefixes, honorifics/titles, adjectives like "World Famous").
    - Accept unambiguous aliases or minor name variants for the same entity (e.g., nicknames vs full names, with/without middle names, common alternate forms) when they clearly refer to the same entity and do not introduce a different one.
    - Numbers/dates: treat words vs digits and digit-grouping (e.g., 1,840 vs 1840) as equivalent. If the expected value is a year, a full date containing that year is acceptable. If the expected value includes month+year, month alone is incomplete.
    - Approximation: If the question or Correct Answer signals approximation (about/approximately/around/over), allow consistent approximate or inequality phrasing near the value.

3) Partial overlap -> partially_correct:
    - At least one expected answer value appears, but any required value is missing and/or at least one value is incorrect/contradictory.
    - The output adds extra distinct answer values for a slot (e.g., listing multiple roles like "actor and director", multiple candidate entities with and/or/slashes/lists) beyond what the Correct Answer expects.

4) Mismatch -> incorrect:
    - None of the expected answer values appear in the System Output, or any provided value contradicts the Correct Answer.

**Definitions and matching guidance:**
- Answer value: an atomic item such as an entity (ID or name), number, date/time, or yes/no.
- Entity matching: give credit only when the expected entity (name/ID or a clear alias/variant) is explicitly present in the System Output. Do not award credit for merely implying the answer without naming it. Do not give credit for broader/narrower/related categories.
- Order: When multiple questions are asked, the correct answers are in that order; the System Output must align to be correct.

[user]
Question: {question}
Correct Answer: {correct_answer}
System Output: {system_output}
\end{lstlisting}

To account for lexical variation in answers and handle verbose expressions (e.g., ``The answer is XX''), we use LLM-as-a-judge to evaluate the correctness of the final answer. The evaluator prompts were tuned semi-automatically on a held-out set of 100 examples per dataset to ensure high agreement with human judgment. See Prompts~\ref{prmt:evaluation-kbqa} and \ref{prmt:evaluation-multiobj-hotpotqa}.

For the GEE analysis, we use the implementation in statsmodels.\footnote{\url{https://www.statsmodels.org/}} To handle repeated trials, we use the question ID for clustered analysis. The formula for the regression is as follows, where $d^*$ and $b^*$ are the normalized depth and breadth of the plan, $x_\text{SH}$ is a binary variable indicating whether the planner is SH.

\noindent \textbf{Atomic KBQA:} \begin{equation}
\text{logit}(P(y=1)) = \beta_0 + \beta_d d^* + \beta_b b^* + \beta_\text{SH} x_\text{SH} + \beta_{d:\text{SH}} (d^* \times x_\text{SH}) + \beta_{b:\text{SH}} (b^* \times x_\text{SH}) + \underbrace{\sum \text{Dataset}_i + \sum \text{LastStep}_j}_{\text{Fixed Effects}}
\end{equation} where C(dataset) is a dummy variable for the dataset (GrailQA, WebQSP, GraphQ), $\text{LastStep}$ is a dummy variable for the type of the last step in the plan.

\noindent \textbf{KQA Pro:} \begin{equation}
\text{logit}(P(y=1)) = \beta_0 + \beta_d d^* + \beta_b b^* + \beta_\text{SH} x_\text{SH} + \beta_{d:\text{SH}} (d^* \times x_\text{SH}) + \beta_{b:\text{SH}} (b^* \times x_\text{SH}) + \underbrace{\sum \text{LastStep}_i}_{\text{Fixed Effects}}
\end{equation} where $\text{LastStep}$ is a dummy variable for the type of the last step in the plan.

\noindent \textbf{Multi-objective HotpotQA:} \begin{equation}
\text{logit}(P(y=1)) = \beta_0 + \beta_d d^* + \beta_b b^* + \beta_\text{SH} x_\text{SH} + \beta_{d:\text{SH}} (d^* \times x_\text{SH}) + \beta_{b:\text{SH}} (b^* \times x_\text{SH}) + \underbrace{\beta_\text{has\_bridge}x_\text{has\_bridge} + \beta_\text{has\_comparison}x_\text{has\_comparison}}_{\text{Fixed Effects}}
\end{equation} where $x_\text{has\_bridge}$ and $x_\text{has\_comparison}$ are binary variables indicating whether the question contains a bridge or comparison sub-question, respectively.

\subsection{External Resources}

\begin{table}[t]
\small
\centering
\caption{A list of pretrained language models used in this study.}
\label{tab:resources-llm}
\begin{tabular}{lcll}
\toprule
\textbf{Name} & \textbf{Parameters} & \textbf{Hosted by} & \textbf{License} \\ \midrule
gpt-4.1-mini-2025-04-14 & n/a & OpenAI & OpenAI Terms of Service \\
gpt-5-mini-2025-08-07 \cite{singh-etal-2025-openai} & n/a & OpenAI & OpenAI Terms of Service \\
gpt-5.2-2025-12-11 \cite{openai-2025-update} & n/a & OpenAI & OpenAI Terms of Service \\
qwen3-235b-a22b-instruct-2507 \cite{yang-etal-2025-qwen3} & 235B & Fireworks AI & Apache 2.0 \\
gemini-3-flash-preview \cite{google-2025-gemini} & n/a & Google Cloud & Google Cloud Terms of Service \\ \bottomrule
\end{tabular}
\end{table}

\begin{table}[t]
\small
\centering
\caption{A list of datasets used in this study.}
\label{tab:resources-data}
\begin{tabular}{lll}
\toprule
\textbf{Name} & \textbf{URL} & \textbf{License} \\
\midrule
KQA Pro~\cite{cao-etal-2022-kqa} & \href{https://huggingface.co/datasets/drt/kqa_pro}{\url{huggingface.co/datasets/drt/kqa_pro}} & MIT \\
GrailQA~\cite{gu-etal-2021-beyond} & \href{https://dki-lab.github.io/GrailQA}{\url{dki-lab.github.io/GrailQA}} & CC BY-SA 4.0 \\
WebQSP~\cite{yih-etal-2016-value} & \href{https://www.microsoft.com/en-us/download/details.aspx?id=52763}{\url{www.microsoft.com/en-us/download/details.aspx?id=52763}} & unspecified \\
GraphQ~\cite{su-etal-2016-generating} & \href{https://github.com/dki-lab/GrailQA/tree/main/data}{\url{github.com/dki-lab/GrailQA/tree/main/data}} & CC BY-SA 4.0 \\
HotpotQA~\cite{yang-etal-2018-hotpotqa} & \href{https://hotpotqa.github.io/}{\url{hotpotqa.github.io}} & CC BY-SA 4.0 \\
En Wikipedia Oct 2017 Dump & \href{https://hotpotqa.github.io/wiki-readme.html}{\url{hotpotqa.github.io/wiki-readme.html}} & CC BY-SA 4.0 \\
Freebase & \href{https://github.com/dki-lab/Freebase-Setup}{\url{github.com/dki-lab/Freebase-Setup}} & CC BY 2.5 \\
\bottomrule
\end{tabular}
\end{table}

\begin{table}[t]
\small
\centering
\caption{A list of software used in this study.}
\label{resources-software}
\begin{tabular}{lll}
\toprule
\textbf{Name} & \textbf{URL} & \textbf{License} \\
\midrule
KoPL~\cite{cao-etal-2022-kqa} & \href{https://github.com/THU-KEG/KoPL}{\url{github.com/THU-KEG/KoPL}} & MIT \\
KBQA-o1~\cite{luo-etal-2025-kbqa} & \href{https://github.com/LHRLAB/KBQA-o1}{\url{github.com/LHRLAB/KBQA-o1}} & MIT \\
Freebase-Setup & \href{https://github.com/dki-lab/Freebase-Setup}{\url{github.com/dki-lab/Freebase-Setup}} & CC0 1.0 \\
openai-python & \href{https://github.com/openai/openai-python}{\url{github.com/openai/openai-python}} & Apache 2.0 \\
Sentence Transformers & \href{https://sbert.net/}{\url{sbert.net}} & Apache 2.0 \\
BGE Embedding Model & \href{https://huggingface.co/BAAI/bge-base-en-v1.5}{\url{huggingface.co/BAAI/bge-base-en-v1.5}} & MIT License \\
statsmodels~\cite{seabold-and-perktold-2010-statsmodels} & \href{https://www.statsmodels.org}{\url{www.statsmodels.org}} & Modified BSD-3-Clause \\
hydra~\cite{yadan-2019-hydra} & \href{https://hydra.cc}{\url{hydra.cc}} & MIT \\
\bottomrule
\end{tabular}
\end{table}

Tables~\ref{tab:resources-llm}, \ref{tab:resources-data} and \ref{resources-software} list the key external resources on which this study relies. Our use of these resources complies with their respective terms of use.

\section{Additional Experimental Results}

\subsection{Detailed GEE Coefficients}\label{app:gee-results}

\begin{table}[t]
\caption{GEE coefficients for KQA Pro experiments (1/2)}
\label{tab:gee-coeff-atomic-kqa-pro-1}
\centering
\begin{tabular}{llccc}
\toprule
\textbf{Model} & \textbf{Feature} & \multicolumn{1}{c}{\textbf{Coefficient}} & \multicolumn{1}{c}{\textbf{$p$}} & \multicolumn{1}{c}{\textbf{Std Err}} \\
\midrule
\multirow[t]{17}{*}{GPT-4.1-mini} & $\beta_0$ & 3.466 & 0.000 & 0.684 \\
 & $\beta_d$ & -0.406 & 0.000 & 0.080 \\
 & $\beta_b$ & -0.723 & 0.002 & 0.232 \\
 & $\beta_\text{SH}$ & 1.542 & 0.001 & 0.464 \\
 & $\beta_{d:\text{SH}}$ & -0.150 & 0.005 & 0.053 \\
 & $\beta_{b:\text{SH}}$ & -0.484 & 0.002 & 0.156 \\
 & $\text{LastStep}$[T.QueryAttr] & 1.661 & 0.006 & 0.600 \\
 & $\text{LastStep}$[T.QueryAttrQualifier] & 0.503 & 0.186 & 0.380 \\
 & $\text{LastStep}$[T.QueryName] & 1.252 & 0.000 & 0.189 \\
 & $\text{LastStep}$[T.QueryRelation] & 1.841 & 0.000 & 0.370 \\
 & $\text{LastStep}$[T.QueryRelationQualifier] & 0.777 & 0.018 & 0.327 \\
 & $\text{LastStep}$[T.SelectAmong] & 0.455 & 0.375 & 0.512 \\
 & $\text{LastStep}$[T.SelectBetween] & 2.323 & 0.000 & 0.416 \\
 & $\text{LastStep}$[T.VerifyDate] & 23.898 & 0.000 & 0.731 \\
 & $\text{LastStep}$[T.VerifyNum] & 0.188 & 0.785 & 0.690 \\
 & $\text{LastStep}$[T.VerifyStr] & 0.806 & 0.014 & 0.329 \\
 & $\text{LastStep}$[T.VerifyYear] & 1.152 & 0.015 & 0.475 \\
\multirow[t]{17}{*}{GPT-5-mini} & $\beta_0$ & 4.202 & 0.000 & 0.959 \\
 & $\beta_d$ & -0.312 & 0.003 & 0.105 \\
 & $\beta_b$ & -0.792 & 0.014 & 0.322 \\
 & $\beta_\text{SH}$ & -0.275 & 0.501 & 0.408 \\
 & $\beta_{d:\text{SH}}$ & 0.030 & 0.584 & 0.054 \\
 & $\beta_{b:\text{SH}}$ & 0.156 & 0.327 & 0.159 \\
 & $\text{LastStep}$[T.QueryAttr] & 1.207 & 0.057 & 0.633 \\
 & $\text{LastStep}$[T.QueryAttrQualifier] & 0.105 & 0.822 & 0.466 \\
 & $\text{LastStep}$[T.QueryName] & 1.274 & 0.000 & 0.286 \\
 & $\text{LastStep}$[T.QueryRelation] & 2.832 & 0.001 & 0.865 \\
 & $\text{LastStep}$[T.QueryRelationQualifier] & 0.513 & 0.231 & 0.428 \\
 & $\text{LastStep}$[T.SelectAmong] & -0.045 & 0.944 & 0.635 \\
 & $\text{LastStep}$[T.SelectBetween] & 2.357 & 0.001 & 0.708 \\
 & $\text{LastStep}$[T.VerifyDate] & 0.677 & 0.405 & 0.813 \\
 & $\text{LastStep}$[T.VerifyNum] & -0.927 & 0.174 & 0.682 \\
 & $\text{LastStep}$[T.VerifyStr] & 0.392 & 0.383 & 0.449 \\
 & $\text{LastStep}$[T.VerifyYear] & 0.791 & 0.228 & 0.656 \\
 \bottomrule
\end{tabular}
\end{table}

\begin{table}[t]
\caption{GEE coefficients for KQA Pro experiments (2/2)}
\label{tab:gee-coeff-atomic-kqa-pro-2}
\centering
\begin{tabular}{llccc}
\toprule
\textbf{Model} & \textbf{Feature} & \multicolumn{1}{c}{\textbf{Coefficient}} & \multicolumn{1}{c}{\textbf{$p$}} & \multicolumn{1}{c}{\textbf{Std Err}} \\
\midrule
\multirow[t]{17}{*}{Qwen3-235B-A22B} & $\beta_0$ & 3.807 & 0.000 & 0.729 \\
 & $\beta_d$ & -0.431 & 0.000 & 0.085 \\
 & $\beta_b$ & -0.728 & 0.004 & 0.251 \\
 & $\beta_\text{SH}$ & -0.852 & 0.040 & 0.415 \\
 & $\beta_{d:\text{SH}}$ & 0.007 & 0.888 & 0.051 \\
 & $\beta_{b:\text{SH}}$ & 0.514 & 0.001 & 0.161 \\
 & $\text{LastStep}$[T.QueryAttr] & 1.272 & 0.021 & 0.550 \\
 & $\text{LastStep}$[T.QueryAttrQualifier] & 0.390 & 0.310 & 0.384 \\
 & $\text{LastStep}$[T.QueryName] & 1.081 & 0.000 & 0.201 \\
 & $\text{LastStep}$[T.QueryRelation] & 2.302 & 0.000 & 0.568 \\
 & $\text{LastStep}$[T.QueryRelationQualifier] & 0.683 & 0.054 & 0.354 \\
 & $\text{LastStep}$[T.SelectAmong] & 1.415 & 0.030 & 0.654 \\
 & $\text{LastStep}$[T.SelectBetween] & 2.459 & 0.000 & 0.583 \\
 & $\text{LastStep}$[T.VerifyDate] & 23.890 & 0.000 & 0.736 \\
 & $\text{LastStep}$[T.VerifyNum] & -0.377 & 0.556 & 0.639 \\
 & $\text{LastStep}$[T.VerifyStr] & 0.879 & 0.032 & 0.409 \\
 & $\text{LastStep}$[T.VerifyYear] & 3.237 & 0.000 & 0.729 \\
\multirow[t]{17}{*}{Gemini-3-Flash} & $\beta_0$ & 3.577 & 0.000 & 0.987 \\
 & $\beta_d$ & -0.249 & 0.030 & 0.115 \\
 & $\beta_b$ & -0.795 & 0.019 & 0.338 \\
 & $\beta_\text{SH}$ & -0.399 & 0.507 & 0.602 \\
 & $\beta_{d:\text{SH}}$ & -0.070 & 0.345 & 0.074 \\
 & $\beta_{b:\text{SH}}$ & 0.115 & 0.571 & 0.203 \\
 & $\text{LastStep}$[T.QueryAttr] & 2.709 & 0.010 & 1.051 \\
 & $\text{LastStep}$[T.QueryAttrQualifier] & 0.805 & 0.046 & 0.403 \\
 & $\text{LastStep}$[T.QueryName] & 2.430 & 0.000 & 0.335 \\
 & $\text{LastStep}$[T.QueryRelation] & 3.694 & 0.000 & 0.955 \\
 & $\text{LastStep}$[T.QueryRelationQualifier] & 1.216 & 0.007 & 0.451 \\
 & $\text{LastStep}$[T.SelectAmong] & 1.948 & 0.065 & 1.054 \\
 & $\text{LastStep}$[T.SelectBetween] & 2.689 & 0.000 & 0.592 \\
 & $\text{LastStep}$[T.VerifyDate] & 23.526 & 0.000 & 0.745 \\
 & $\text{LastStep}$[T.VerifyNum] & -0.242 & 0.737 & 0.721 \\
 & $\text{LastStep}$[T.VerifyStr] & 1.600 & 0.003 & 0.542 \\
 & $\text{LastStep}$[T.VerifyYear] & 2.410 & 0.001 & 0.738 \\
 \bottomrule
\end{tabular}
\end{table}

\begin{table}[t]
\caption{GEE coefficients for Atomic KBQA experiments}
\label{tab:gee-coeff-atomic-kbqa}
\begin{tabular}{llccc}
\toprule
\textbf{Model} & \textbf{Feature} & \multicolumn{1}{c}{\textbf{Coefficient}} & \multicolumn{1}{c}{\textbf{$p$}} & \multicolumn{1}{c}{\textbf{Std Err}} \\
\midrule
\multirow[t]{11}{*}{GPT-4.1-mini}  & $\text{Dataset}$[T.graphq] & -0.819 & 0.000 & 0.143 \\
 & $\beta_d$ & -0.330 & 0.001 & 0.097 \\
 & $\beta_b$ & -0.269 & 0.017 & 0.113 \\
 & $\beta_\text{SH}$ & 0.474 & 0.059 & 0.250 \\
 & $\beta_{d:\text{SH}}$ & -0.100 & 0.103 & 0.062 \\
 & $\beta_{b:\text{SH}}$ & -0.130 & 0.114 & 0.082 \\
 & $\text{Dataset}$[T.webqsp] & -0.386 & 0.100 & 0.235 \\
 & $\text{LastStep}$[T.find\_relation] & -0.284 & 0.414 & 0.348 \\
 & $\text{LastStep}$[T.merge] & -1.000 & 0.000 & 0.243 \\
 & $\text{LastStep}$[T.order] & -0.268 & 0.416 & 0.330 \\
\multirow[t]{11}{*}{GPT-5-mini} & $\beta_0$ & 3.350 & 0.000 & 0.495 \\
 & $\beta_d$ & -0.350 & 0.000 & 0.097 \\
 & $\beta_b$ & -0.383 & 0.002 & 0.124 \\
 & $\beta_\text{SH}$ & -0.211 & 0.294 & 0.201 \\
 & $\beta_{d:\text{SH}}$ & 0.031 & 0.548 & 0.052 \\
 & $\beta_{b:\text{SH}}$ & 0.029 & 0.684 & 0.070 \\
 & $\text{Dataset}$[T.graphq] & -0.542 & 0.001 & 0.161 \\
 & $\text{Dataset}$[T.webqsp] & -0.491 & 0.053 & 0.253 \\
 & $\text{LastStep}$[T.find\_relation] & -0.051 & 0.895 & 0.383 \\
 & $\text{LastStep}$[T.merge] & -0.325 & 0.260 & 0.288 \\
 & $\text{LastStep}$[T.order] & -0.481 & 0.172 & 0.352 \\
\multirow[t]{11}{*}{Qwen3-235B-A22B} & $\beta_0$ & 3.459 & 0.000 & 0.473 \\
 & $\beta_d$ & -0.325 & 0.002 & 0.103 \\
 & $\beta_b$ & -0.531 & 0.000 & 0.118 \\
 & $\beta_{b:\text{SH}}$ & 0.060 & 0.416 & 0.074 \\
 & $\beta_\text{SH}$ & -0.080 & 0.724 & 0.227 \\
 & $\beta_{d:\text{SH}}$ & -0.016 & 0.794 & 0.061 \\
 & $\text{Dataset}$[T.graphq] & -0.611 & 0.000 & 0.150 \\
 & $\text{Dataset}$[T.webqsp] & -0.176 & 0.495 & 0.258 \\
 & $\text{LastStep}$[T.find\_relation] & -0.162 & 0.650 & 0.358 \\
 & $\text{LastStep}$[T.merge] & -0.662 & 0.005 & 0.238 \\
 & $\text{LastStep}$[T.order] & -0.021 & 0.951 & 0.334 \\
\multirow[t]{11}{*}{Gemini-3-Flash} & $\beta_0$ & 3.144 & 0.000 & 0.560 \\
 & $\beta_d$ & -0.349 & 0.006 & 0.126 \\
 & $\beta_b$ & -0.235 & 0.098 & 0.142 \\
 & $\beta_\text{SH}$ & 0.150 & 0.601 & 0.287 \\
 & $\beta_{d:\text{SH}}$ & -0.007 & 0.920 & 0.065 \\
 & $\beta_{b:\text{SH}}$ & -0.331 & 0.001 & 0.102 \\
 & $\text{Dataset}$[T.graphq] & -0.556 & 0.000 & 0.160 \\
 & $\text{Dataset}$[T.webqsp] & -0.126 & 0.656 & 0.283 \\
 & $\text{LastStep}$[T.find\_relation] & 0.000 & 1.000 & 0.397 \\
 & $\text{LastStep}$[T.merge] & -0.402 & 0.136 & 0.270 \\
 & $\text{LastStep}$[T.order] & 0.021 & 0.956 & 0.371 \\
 \bottomrule
\end{tabular}
\end{table}

\begin{table}[t]
\caption{GEE coefficients for multi-objective HotpotQA experiments}
\label{tab:gee-coeff-hotpot-qa}
\begin{tabular}{llccc}
\toprule
\textbf{Model} & \textbf{Feature} & \multicolumn{1}{c}{\textbf{Coefficient}} & \multicolumn{1}{c}{\textbf{$p$}} & \multicolumn{1}{c}{\textbf{Std Err}} \\
\midrule
\multirow[t]{8}{*}{GPT-4.1-mini} & $\beta_0$ & 2.459 & 0.000 & 0.298 \\
 & $\beta_d$ & -0.781 & 0.000 & 0.182 \\
 & $\beta_b$ & -0.155 & 0.000 & 0.019 \\
 & $\beta_\text{SH}$ & 0.256 & 0.227 & 0.212 \\
 & $\beta_{d:\text{SH}}$ & -0.006 & 0.967 & 0.141 \\
 & $\beta_{b:\text{SH}}$ & -0.014 & 0.312 & 0.014 \\
 & $\beta_\text{has\_bridge}$ & -0.735 & 0.000 & 0.143 \\
 & $\beta_\text{has\_comparison}$ & 0.529 & 0.000 & 0.143 \\
\multirow[t]{8}{*}{GPT-5-mini} & $\beta_0$ & 2.376 & 0.000 & 0.270 \\
 & $\beta_d$ & -0.839 & 0.000 & 0.168 \\
 & $\beta_b$ & -0.127 & 0.000 & 0.018 \\
 & $\beta_\text{SH}$ & -0.235 & 0.272 & 0.214 \\
 & $\beta_{d:\text{SH}}$ & 0.102 & 0.456 & 0.137 \\
 & $\beta_{b:\text{SH}}$ & 0.009 & 0.525 & 0.014 \\
 & $\beta_\text{has\_bridge}$ & -0.638 & 0.000 & 0.131 \\
 & $\beta_\text{has\_comparison}$ & 0.543 & 0.000 & 0.130 \\
\multirow[t]{8}{*}{Qwen3-235B-A22B} & $\beta_0$ & 1.730 & 0.000 & 0.271 \\
 & $\beta_d$ & -0.480 & 0.004 & 0.169 \\
 & $\beta_b$ & -0.115 & 0.000 & 0.018 \\
 & $\beta_\text{SH}$ & 0.933 & 0.000 & 0.243 \\
 & $\beta_{d:\text{SH}}$ & -0.213 & 0.164 & 0.153 \\
 & $\beta_{b:\text{SH}}$ & -0.033 & 0.028 & 0.015 \\
 & $\beta_\text{has\_bridge}$ & -0.868 & 0.000 & 0.141 \\
 & $\beta_\text{has\_comparison}$ & 0.452 & 0.000 & 0.130 \\
\multirow[t]{8}{*}{Gemini-3-Flash} & $\beta_0$ & 1.659 & 0.000 & 0.283 \\
 & $\beta_d$ & -0.546 & 0.003 & 0.182 \\
 & $\beta_b$ & -0.090 & 0.000 & 0.018 \\
 & $\beta_\text{SH}$ & 0.850 & 0.003 & 0.288 \\
 & $\beta_{d:\text{SH}}$ & -0.434 & 0.024 & 0.192 \\
 & $\beta_{b:\text{SH}}$ & -0.066 & 0.000 & 0.017 \\
 & $\beta_\text{has\_bridge}$ & -0.609 & 0.000 & 0.138 \\
 & $\beta_\text{has\_comparison}$ & 0.313 & 0.019 & 0.133 \\
 \bottomrule
\end{tabular}
\end{table}

Tables \ref{tab:gee-coeff-atomic-kqa-pro-1}, \ref{tab:gee-coeff-atomic-kqa-pro-2}, \ref{tab:gee-coeff-atomic-kbqa}, and \ref{tab:gee-coeff-hotpot-qa} show all the GEE coefficients.

\subsection{Case Study on Repetitive Tool Calls}\label{app:case-study}

\begin{table}[t]
    \centering
    \caption{SH with Gemini-3-Flash under low-robustness retrieval (top-1) on multi-objective HotpotQA ($k=2$). This table focuses on one sub-question: ``Which director, John Schlesinger or Barbara Albert, was also a writer and film producer?''}
    \label{tab:case-study-full-sh}
    \begin{tabular}{rlp{0.7\linewidth}}
    \toprule
    \multicolumn{1}{l}{\multirow[t]{3}{*}{Step 1}} & Reasoning & I'm currently engaged in evaluating both John Schlesinger and Barbara Albert, determining if each functioned as a writer and producer. After gathering the facts, a comparison will readily reveal the answer. Parallel to this, I'm setting up searches to uncover the release companies behind Cinderella and The Emperor's New Groove to formulate my response regarding the latter question. \\
    & Tool & > search(Was John Schlesinger a writer and film producer?) \\
    & Output & \begin{tabular}[c]{p{1\linewidth}}Error in search: Failed to find the answer to "Was John Schlesinger a writer and film producer?"\\ No supporting information found in the search result: information is about Richard Schlesinger, not John Schlesinger.\\ Retry with a different question or try a different tool.\end{tabular} \\
    \midrule
    \multirow[t]{3}{*}{Step 2} & Reasoning & I'm starting by investigating the first question. I've initiated Step 0: searching for John Schlesinger's roles. The plan is to establish if he wrote or produced films, as well as directed. After that, I will move on to step 1 and do the same for Barbara Albert. \\
    & Tool & > search(What were the occupations of director John Schlesinger?) \\
    & Output & \begin{tabular}[c]{p{1\linewidth}}Error in search: Failed to find the answer to "Was John Schlesinger a writer and film producer?"\\ No supporting information found in the search result: information is about Richard Schlesinger, not John Schlesinger.\\ Retry with a different question or try a different tool.\end{tabular} \\
    \midrule
    \multirow[t]{3}{*}{Step 3} & Reasoning & I'm now formulating my search queries. I'll begin by querying about the occupations of John Schlesinger and Barbara Albert. Following that, I'll identify the release companies for "Cinderella" and "The Emperor's New Groove." This information will then allow me to address the initial questions posed. I'm focusing on efficiency and clarity to ensure accurate results. \\
    & Tool & > search(What were the occupations of John Schlesinger?) \\
    & Output & \begin{tabular}[c]{p{1\linewidth}}Error in search: Failed to find the answer to "Was John Schlesinger a writer and film producer?"\\ No supporting information found in the search result: information is about Richard Schlesinger, not John Schlesinger.\\ Retry with a different question or try a different tool.\end{tabular} \\
    \midrule
    \multirow[t]{3}{*}{Step 4} & Reasoning & I've begun looking into the roles of John Schlesinger and Barbara Albert. Initial searches for their occupations are underway. I'll need to sift through the results to see if either one of them was both a writer and a film producer. The plan is to compare the findings to pinpoint the director who fulfills both criteria. I'll then move on to the second question. \\
    & Tool & > search(What were the occupations of John Schlesinger?) \\
    & Output & \begin{tabular}[c]{p{1\linewidth}}Error in search: Failed to find the answer to "Was John Schlesinger a writer and film producer?"\\ No supporting information found in the search result: information is about Richard Schlesinger, not John Schlesinger.\\ Retry with a different question or try a different tool.\end{tabular} \\
    \midrule
    \multirow[t]{3}{*}{Step 5} & Reasoning & I'm now starting to investigate John Schlesinger. My initial step is to determine if he also worked as a writer and film producer. Then, I'll turn my attention to Barbara Albert, following the same procedure. After that, I will compare the data to formulate a response to the user's first question. \\
    & Tool & > search(What were the occupations of director John Schlesinger?) \\
    & Output & \begin{tabular}[c]{p{1\linewidth}}Error in search: Failed to find the answer to "Was John Schlesinger a writer and film producer?"\\ No supporting information found in the search result: information is about Richard Schlesinger, not John Schlesinger.\\ Retry with a different question or try a different tool.\end{tabular} \\
    \bottomrule
    \end{tabular}
\end{table}

\begin{table}[t]
    \centering
    \caption{FH with Gemini-3-Flash under low-robustness retrieval (top-1) on multi-objective HotpotQA ($k=2$). This table focuses on one sub-question: ``Which director, John Schlesinger or Barbara Albert, was also a writer and film producer?''}
    \label{tab:case-study-full-fh}
    \begin{tabular}{rlp{0.7\linewidth}}
    \toprule
\multirow[t]{3}{*}{Initial} & Reasoning & I've initiated searches to determine if John Schlesinger and Barbara Albert were both writers and producers, as planned. Once I have those details, a comparison will be conducted to pinpoint the correct individual. I'm also preparing to find the release companies for "Cinderella" and "The Emperor's New Groove" to compare them. \\
 & Plan & > [\$0 = search(Was John Schlesinger a writer and film producer?), \$1 = search(Was Barbara Albert a writer and film producer?), \$2 = reasoning(John Schlesinger is described as \$0, and Barbara Albert is described as \$1; based on which one was also a writer and film producer, output the correct person.), ...] \\
 & Output & \begin{tabular}[c]{p{1\linewidth}}Error in search \$0: Failed to find the answer to "Was John Schlesinger a writer and film producer?"\\ No supporting information found in the search result: information is about Richard Schlesinger, not John Schlesinger.\\ Retry with a different question or try a different tool.\end{tabular} \\
 \midrule
\multirow[t]{3}{*}{Replanning 1} & Reasoning & I've restarted my search for John Schlesinger's occupations, focusing on director, writer, and producer roles, since my initial attempt didn't yield the right information. Also, I'm planning to look into Barbara Albert's roles. My next step will be to compare the findings to pinpoint the director who was also a writer and film producer. \\
 & Plan & > [\$1 = search(What were the occupations of director John Schlesinger?), \$2 = search(What were the occupations of director Barbara Albert?), ...] \\
 & Output & \begin{tabular}[c]{p{1\linewidth}}Error in search \$1: Failed to find the answer to "Was John Schlesinger a writer and film producer?"\\ No supporting information found in the search result: information is about Richard Schlesinger, not John Schlesinger.\\ Retry with a different question or try a different tool.\end{tabular} \\
 \midrule
\multirow[t]{3}{*}{Replanning 2} & Reasoning & I'm now focusing on Barbara Albert's occupations, and then refining how I search for John Schlesinger's roles. I plan to query directly whether Schlesinger worked as a writer and producer. After that, I'll compare the findings to answer the first question. For the second question, I will pinpoint the release companies for both films, and compare the results. \\
 & Plan & > [\$2 = search(What are the occupations of Barbara Albert (director)?), \$3 = search(Was John Schlesinger a writer and film producer in addition to being a director?), ...] \\
 & Output & \begin{tabular}[c]{p{1\linewidth}}Error in search \$3: Failed to find the answer to "Was John Schlesinger a writer and film producer?"\\ No supporting information found in the search result: information is about Richard Schlesinger, not John Schlesinger.\\ Retry with a different question or try a different tool.\end{tabular} \\
 \midrule
\multirow[t]{2}{*}{Replanning 3} & Reasoning & I'm focusing on John Schlesinger's occupations, trying to confirm if he was also a writer or producer. I'll use a more specific search term this time around. If that doesn't work, I'll leverage Barbara Albert's information to answer the first question. After that, I'll tackle the release companies. \\
 & Plan & > [\$4 = search(What were the professional roles of the English film director John Schlesinger?), \$5 = reasoning(Based on the occupations of Barbara Albert (\$2) and John Schlesinger (\$4), which one was also a writer and film producer?), ...] \\
    \bottomrule
    \end{tabular}
\end{table}

Tables \ref{tab:case-study-full-sh} and \ref{tab:case-study-full-fh} show the full trajectories corresponding to the case study examples in Table \ref{tab:repetition-case-study} (\S\ref{sec:exp-failure-mode}).

\end{document}